\documentclass[sigconf]{acmart}
\usepackage{booktabs}
\usepackage{subfig}
\usepackage{amsmath}
\usepackage{dsfont}
\usepackage{balance}

\DeclareMathOperator*{\argmax}{arg\,max}

\AtBeginDocument{%
  }

\copyrightyear{2022}
\acmYear{2022}
\setcopyright{acmcopyright}
\acmConference[KDD '22] {Proceedings of the 28th ACM SIGKDD Conference on Knowledge Discovery and Data Mining}{August 14--18, 2022}{Washington, DC, USA.}
\acmBooktitle{Proceedings of the 28th ACM SIGKDD Conference on Knowledge Discovery and Data Mining (KDD '22), August 14--18, 2022, Washington, DC, USA}
\acmPrice{15.00}
\acmISBN{978-1-4503-9385-0/22/08}
\acmDOI{10.1145/3534678.3539209}

\begin{document}

\title{Duplex Conversation: \\ Towards Human-like Interaction in Spoken Dialogue Systems}

\author{Ting-En Lin}
\orcid{0000-0003-0650-7286}
\affiliation{%
  \institution{Alibaba Group}
  \state{Beijing}
  \country{China}}
\email{ting-en.lte@alibaba-inc.com}

\author{Yuchuan Wu}
\affiliation{%
  \institution{Alibaba Group}
  \state{Beijing}
  \country{China}}

\author{Fei Huang}
\affiliation{%
  \institution{Alibaba Group}
  \state{Beijing}
  \country{China}}

\author{Luo Si}
\affiliation{%
  \institution{Alibaba Group}
  \state{Beijing}
  \country{China}}

\author{Jian Sun}
\affiliation{%
  \institution{Alibaba Group}
  \state{Beijing}
  \country{China}}

\author{Yongbin Li}
\authornote{Yongbin Li is the corresponding author.}
\affiliation{%
  \institution{Alibaba Group}
  \state{Beijing}
  \country{China}}
\email{shuide.lyb@alibaba-inc.com}

\renewcommand{\shortauthors}{Ting-Eng Lin et al.}
\renewcommand{\shorttitle}{Duplex Conversation}

\begin{abstract}
In this paper, we present \textit{Duplex Conversation}, a multi-turn, multimodal spoken dialogue system that enables telephone-based agents to interact with customers like a human. We use the concept of full-duplex in telecommunication to demonstrate what a human-like interactive experience should be and how to achieve smooth turn-taking through three subtasks: user state detection, backchannel selection, and barge-in detection. Besides, we propose semi-supervised learning with multimodal data augmentation to leverage unlabeled data to increase model generalization. Experimental results on three sub-tasks show that the proposed method achieves consistent improvements compared with baselines. We deploy the Duplex Conversation to Alibaba intelligent customer service and share lessons learned in production. Online A/B experiments show that the proposed system can significantly reduce response latency by 50\%.
\end{abstract}




\maketitle

\begin{figure}
\centering
\subfloat[Simplex: One-way communication, and the direction is fixed.]{
    \label{subfig:example_1}
    \includegraphics[width= \linewidth]{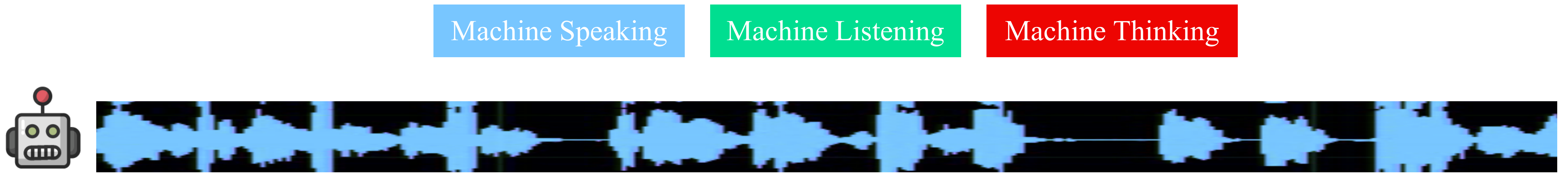} } 
    
\subfloat[Half-duplex: Two-way communication, but not simultaneously.]{
    \label{subfig:example_2}
    \includegraphics[width= 0.96\linewidth]{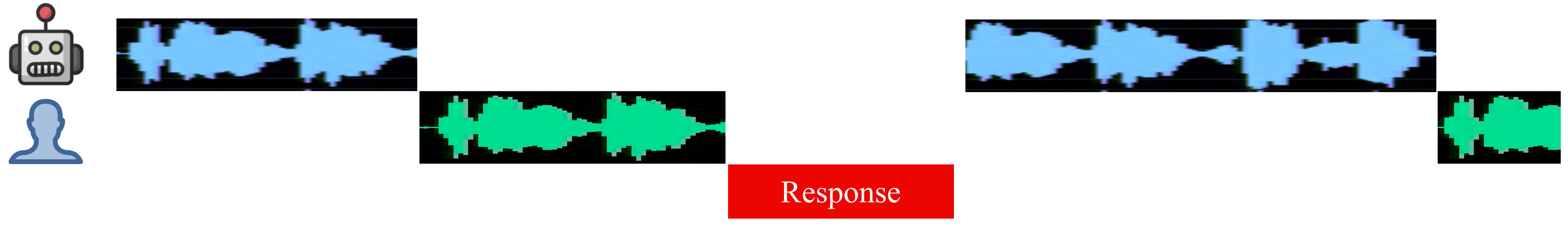} } 

\subfloat[Full-duplex: Two-way communication, simultaneously.]{
    \label{subfig:example_3}
    \includegraphics[width=0.96\linewidth]{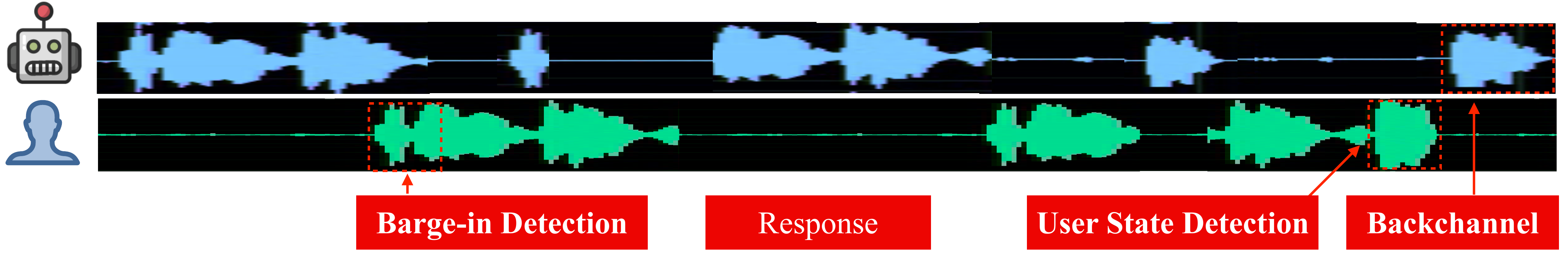}} 
\caption{The illustration of Simplex, Half-Duplex, and Full-Duplex.}
\label{fig:example}
\end{figure}

\section{Introduction}
How to make intelligent service robots interact with people like humans is a critical and difficult challenge in spoken dialogue systems (SDS) \cite{marge2022spoken}. With the rapid development of artificial intelligence, the application of SDS has made significant progress in recent years \cite{wang2021mell, liu2020towards, liu2019automatic, yan2018coupled}. Intelligent assistants, such as Alexa, Siri, Cortana, Google Assistant, and telephone-based intelligent customer service, such as Alibaba intelligent service robots, have entered people's daily life. 
However, most existing commercial SDS only focus on \textbf{what} the agent should say to fulfill user needs, ignoring the importance of \textbf{when} to interact with the user. Rather than waiting for a fixed-length silence after user speech and then triggering a system response, the agent should be able to coordinate who is currently talking and when the next person could start to talk \cite{skantze2021turn}. It is difficult for the agent to make flexible turn-taking with small gaps and overlaps \cite{raux2008flexible} as humans do. 

\begin{figure*}[ht!]
\centering
\subfloat[Regular human-machine conversation.]{
    \label{subfig:ability_1}
    \includegraphics[width= 0.40\linewidth]{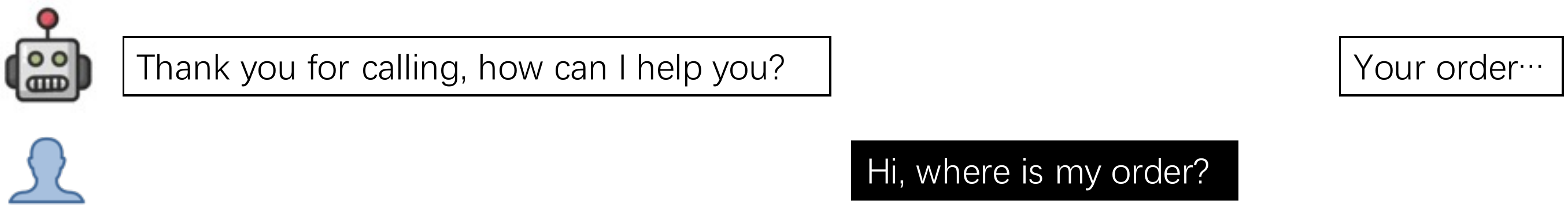}} \hspace{0.2cm}
\subfloat[Machine response with human-like backchannel.]{
    \label{subfig:ability_2}
    \includegraphics[width= 0.55\linewidth]{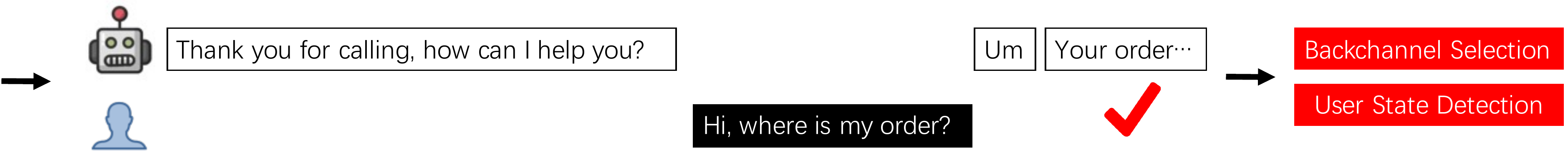}} 
    
\subfloat[Incomplete query caused by hesitant users.]{
    \label{subfig:ability_5}
    \includegraphics[width=0.40\linewidth]{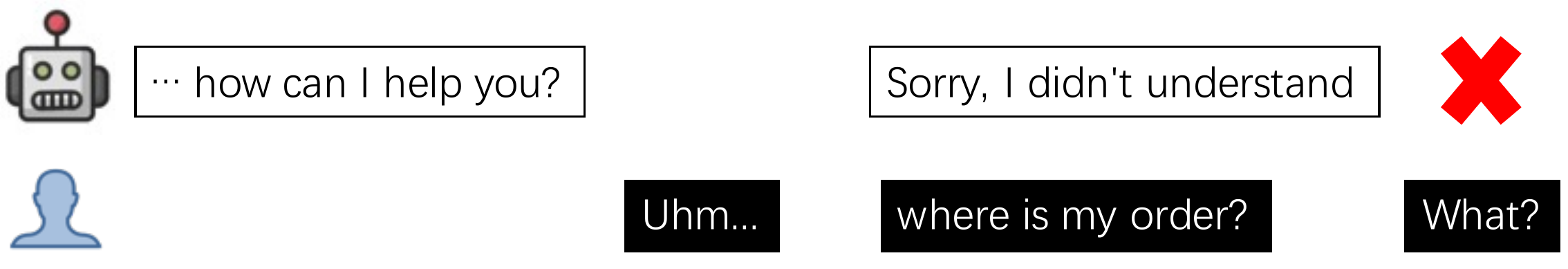}} \hspace{0.2cm}
\subfloat[Detecting hesitant query and guide user to complete it.]{
    \label{subfig:ability_6}
    \includegraphics[width=0.55\linewidth]{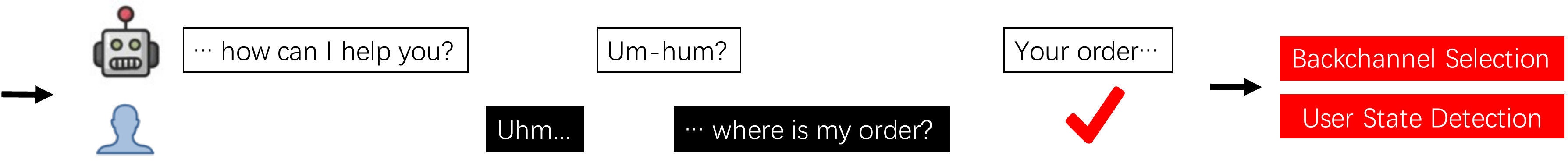}} 

\subfloat[User false barge-in from noise.]{
    \label{subfig:ability_3}
    \includegraphics[width=0.41\linewidth]{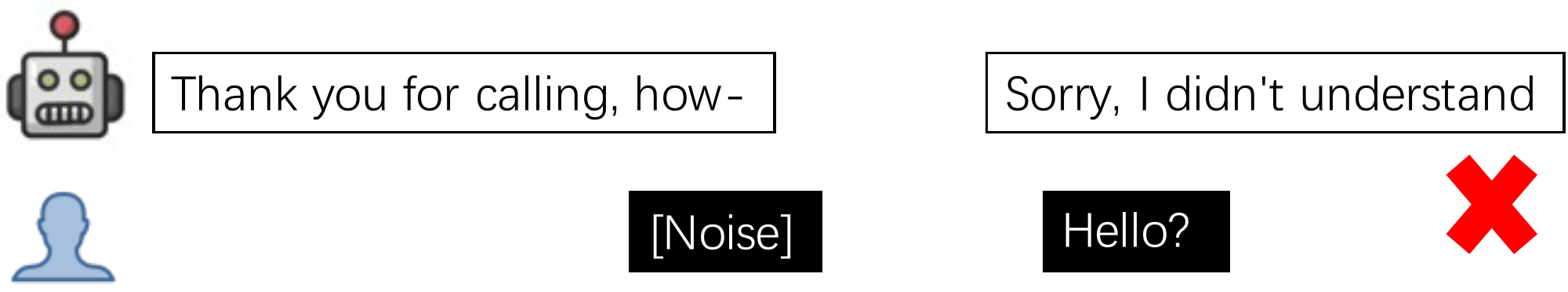}} \hspace{0.2cm}
\subfloat[User barge-in with semantic robustness.]{
    \label{subfig:ability_4}
    \includegraphics[width=0.56\linewidth]{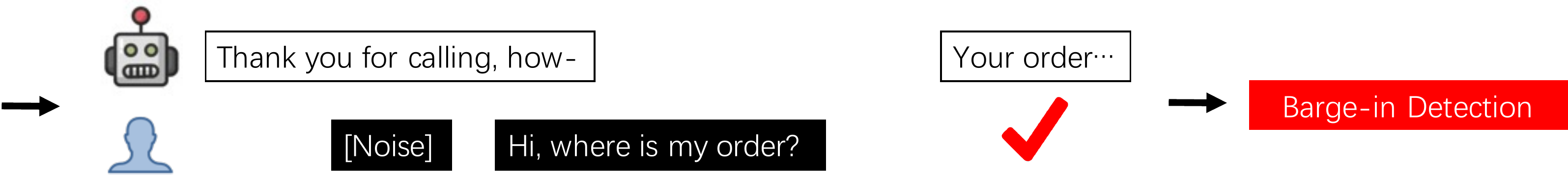}} 
\caption{Examples of problems in spoken dialogue systems and how duplex conversation solves them.}
\label{fig:ability}
\end{figure*}

We use the concepts of simplex, half-duplex, and full-duplex in telecommunication to better illustrate what flexible turn-taking is, as shown in Figure \ref{fig:example}. First, the simplest form of communication is simplex. In this situation, the sender and receiver are fixed, and the communication is unidirectional, such as TV or radio broadcasting. Second, half-duplex allows changing the sender and receiver, and the communication is bidirectional, but not simultaneously. For example, pagers or most telephone-based intelligent customer service fall into this category. Finally, full-duplex allows simultaneous two-way communication without restrictions, such as natural human-to-human interaction. 

In full-duplex, the ultimate goal is to improve the efficiency of communication. For full-duplex in SDS, agents should be able to take turns with customers as smoothly as human-to-human communication by reducing the time when both parties are talking or silent at the same time \cite{sacks1978simplest}. Therefore, the agent should be able to speak, listen, and think simultaneously when necessary, as shown in Figure \ref{subfig:example_3}. It also needs to determine whether it is the right time to insert the backchannel response, or whether the user wants to interrupt the agent or has not finished speaking. By reducing the time when both parties are talking or silent simultaneously, agents should be able to take turns with customers as smoothly as human-to-human communication.

There are several attempts to model full-duplex behavior and apply it to SDS agents in the literature. For example, Google Duplex \cite{leviathan2018google} integrates the backchannel response, such as "\textit{Um-hum, Well...}", into SDS agent for restaurant booking agents and dramatically improves the naturalness. Microsoft Xiaoice \cite{zhou2020design} proposes the rhythm control module to achieve better turn-taking for the intelligent assistant. Jin et al. \cite{jin2021duplex} demonstrate an outbound agent that could recognize user interruptions and discontinuous expressions. Inoue et al. \cite{inoue2020attentive} built an android called ERICA with backchannel responses for attentive listening. 

Nonetheless, there are still room for improvement in the above works. First, most systems remain experimental and only have part of full-duplex capability. Second, most models only consider the transcribed text and ignore audio input when making decisions. It fails to capture acoustic features such as prosody, rhythm, pitch, and intensity, leading to poor results. Finally, there is no large-scale application in the telephone-based intelligent customer service with the ability to generalize across domains. 

In the paper, we propose \textit{Duplex Conversation}, a system that enables SDS to perform flexible turn-taking like humans. Our system has three full-fledged capabilities, including user state detection, backchannel selection, and barge-in detection for the full-duplex experience. Unlike previous systems that only consider the transcribed text, we propose a multimodal model to build turn-taking abilities using audio and text as inputs \cite{faruqui2021revisiting}. By introducing audio into the decision-making process, we can more accurately detect whether the user has completed their turn or detect background noise such as chatter to avoid false interruption. Furthermore, we also propose multimodal data augmentation to improve robustness and use large-scale unlabeled data through semi-supervised learning to improve domain generalization. Experimental results show that our approach achieves consistent improvements compared with baselines. 

\begin{table*}
  \caption{The detail decision-making process for each user state over time.}
  \label{tab:decision_making}
  \begin{tabular}{cccc}
    \toprule
                                & Turn-switch              & Turn-keep               & Turn-keep with hesitation  \\
    \midrule
    IPU threshold (200ms silence)      & Backchannel response & Keep                    & Keep         \\
    VAD threshold (800ms silence)      & End-of-turn          & Keep                    & Backchannel response   \\
    VAD threshold $\sim$Timeout        & -                    & Concatenate ASR results & Concatenate ASR results \\
    Timeout (3000ms silence)           & -                    & End-of-turn             & End-of-turn \\   
    \bottomrule
  \end{tabular}
\end{table*}

We summarize our contribution as follows. First, we present a spoken dialogue system called Duplex Conversation, equipped with three full-fledged capabilities to enable a human-like interactive experience. Second, we model turn-taking behaviors through multimodal models of speech and text, and propose multimodal data augmentation and semi-supervised learning to improve generalization. Experiments show that the proposed method yields significant improvements compared to the baseline. Finally, we deploy the proposed system to Alibaba intelligent customer service and summarize the lessons learned during the deployment. To the best of our knowledge, we are the first to describe such a duplex dialogue system and provide deployment details in telephone-based intelligent customer service.


\section{Duplex Conversation}
In this section, we first outline the three capabilities included in the duplex conversation, with examples to aid understanding. Then, we describe three subtasks in detail: user state detection, backchannel selection, and barge-in detection.

\subsection{Overview}
We show three pairs of duplex conversation examples in Figure \ref{fig:ability}. The first skill is to give robots \textbf{a human-like backchannel response}. For example, Figure \ref{subfig:ability_1} is a regular question-answering process in the spoken dialogue system. We can quickly insert backchannel responses, such as "yeah, um-hum", before the official responds, as shown in Figure \ref{subfig:ability_2}, thereby reducing the response latency experienced by the user. 

The second skill is to \textbf{detect the hesitant query and guide the user to finish it}. In Figure \ref{subfig:ability_5}, due to the user's natural delay or pause in the process of speaking, the user's request is incorrectly segmented by the robot and responded. As shown in Figure \ref{subfig:ability_6}, if the model detects that the user's recent speech has not finished speaking, the robots would use a backchannel response to guide the user to finish their words. After the complete request is obtained by concatenating user utterances, it is sent to the core dialogue engine to improve the dialogue effect.

The third skill is to \textbf{identify user barge-in intent} while rejecting false interruptions from noise. As shown in Figure \ref{subfig:ability_3} , most existing user interruptions are judged based on simple rules, such as ASR confidence, acoustic intensity, etc. However, simple rule-based strategies can easily lead to false interruptions and recognize noises and chatter around the user as normal requests. In Figure \ref{subfig:ability_4}, it demonstrates that the agents should be able to reject noise and handle user barge-in requests correctly.

To achieve the above three capabilities, we build three models and corresponding tasks, including user state detection, backchannel selection, and barge-in detection. Note that user state detection and barge-in detection are multimodal models whose inputs are audio and text, while the input for backchannel selection is text.

\subsection{User State Detection} 
User state detection could be considered an extension of end-of-turn detection, including three user states: turn-switch, turn-keep, and turn-keep with hesitation. We build a multimodal model with audio and text as input to obtain a more accurate classification and design different policies for each state, as shown in Table \ref{tab:decision_making}. 

Traditional spoken dialogue systems only perform inference at the end of the turn. In contrast, an ideal duplex conversation could continuously perform inference while the user speaks. However, continuous inference produces a lot of useless predictions and imposes an enormous burden on system performance. Therefore, We choose a compromise solution and perform inference on Inter-Pausal Units (IPUs) \cite{skantze2021turn}. 

Typically, we determine user turns by the VAD silence threshold. Here, we define the smaller VAD silence threshold as the IPU threshold and cut user utterances into IPUs. If the user state is turn-switch, we will request the backchannel selection module for the proper response. When the silence reaches the VAD threshold, the agent considers it as an end-of-turn and sends the user query to the core dialogue engine.

If the user state is turn-keep, it means the user has not finished speaking, and the agent will not end the user's turn unless the silence duration reaches the timeout. Assuming that the user continues to speak before the timeout, the system will concatenate the ASR transcript and re-monitor the silence duration.

Assuming the user state is turn-keep with hesitation, it will additionally request the backchannel selection module at the IPU threshold on top of turn-keep logic. User requests are often semantically incomplete in such cases, so we let agents insert appropriate backchannel responses to guide users to finish their sentences.

\subsection{Backchannel Selection} 
The backchannel selection module is designed to respond appropriately to the given query and user state. Since the set of suitable backchannels is limited, we mine and count all possible backchannel responses and select the ten most suitable responses from crowdsourcing customer service conversations. 

There may be multiple possible backchannel responses for the same query. We construct training data for multi-label classification by merging identical user queries and queries with the same dialogue acts. We also group similar queries by hierarchical clustering and count the distribution of backchannel responses for each cluster. After normalization, we get the probability distribution of backchannel responses for corresponding clusters, and use it as the soft label in the multi-label classification.

Since the model input relies only on the transcribed text, we choose a simple text convolutional neural network with binary cross-entropy as the classifier and select the one with the highest probability as the response. If there are multiple responses with a probability above the threshold, we choose a response by weighted random selection for better diversity.

\begin{figure}[ht]
  \centering
  \includegraphics[width=0.9\linewidth]{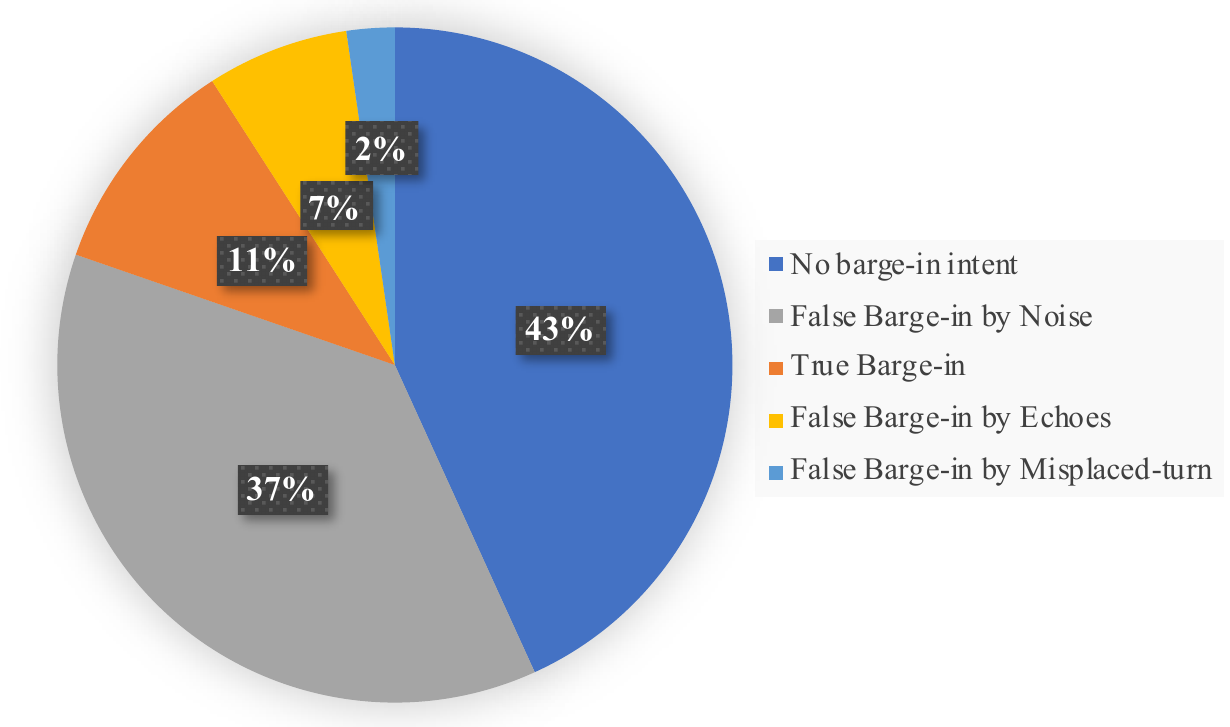}
  \caption{The analysis of rule-based barge-in detection.}
  \label{fig:barge-in-analysis}  
\end{figure}

\begin{figure*}
  \centering
  \includegraphics[width=0.985\linewidth]{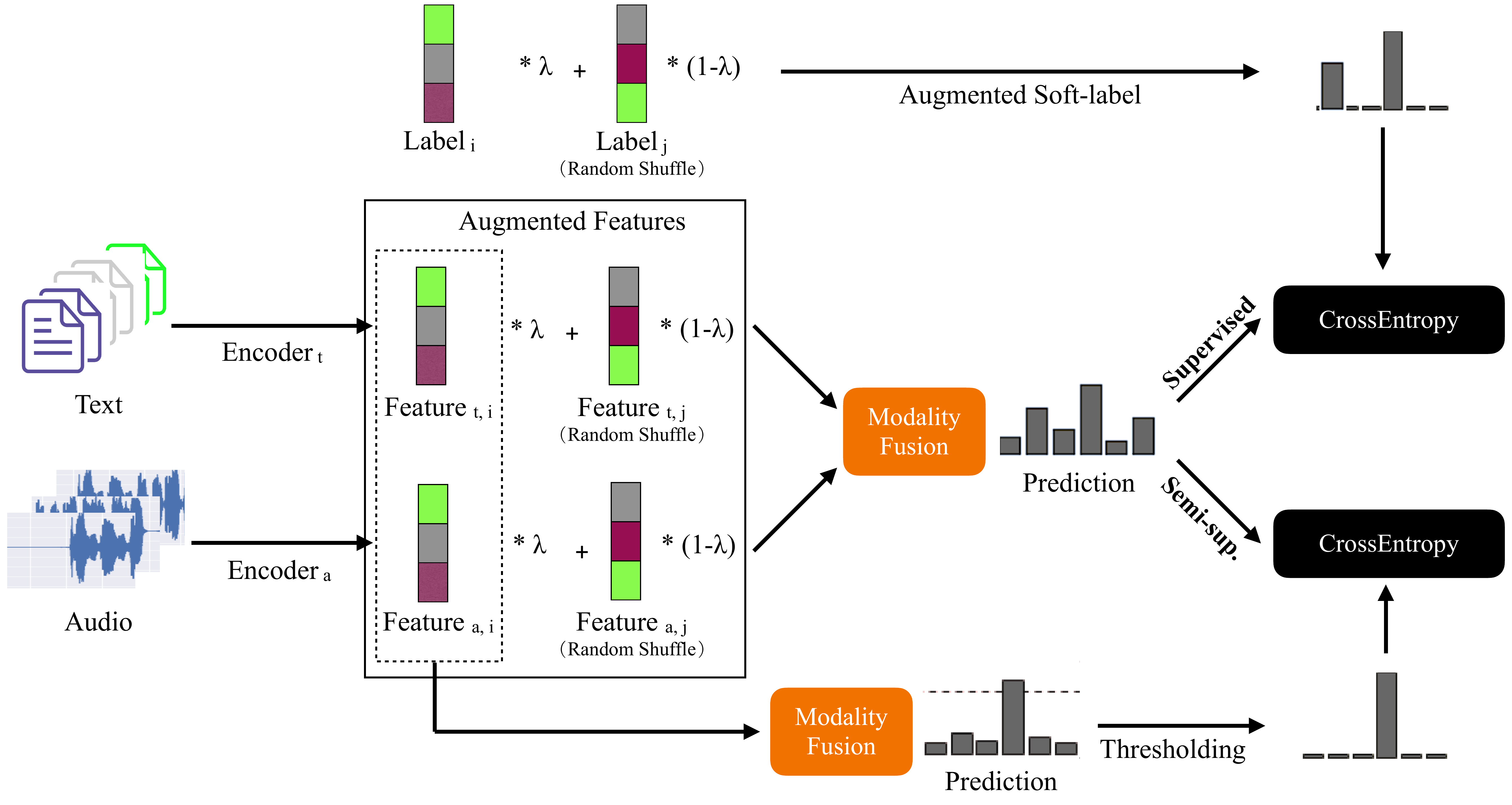}
  \caption{The proposed semi-supervised classifier with multimodal data augmentation. }
  \label{fig:model}  
\end{figure*}

\subsection{Barge-in Detection} 
Allowing the user to interrupt while the robot is speaking is an essential feature of duplex conversations. However, simple rule-based policies often lead to many false interruptions, resulting in a poor experience.

We conduct an empirical analysis of user barge-in behavior in historical traffic with the rule-based policy. If the streaming ASR has intermediate results and the recognition confidence is above the threshold, we consider that the user wants to interrupt the agent. As shown in Figure \ref{fig:barge-in-analysis}, only 11\% are user barge-in requests and the rest 89\% are all false interruptions. Therefore, how to reduce false interruptions becomes the core of the problem.

We divide false barge-in into four categories: (1) User has no intention to interrupt, such as greeting or backchannel (2) Noises, such as environment noises and background vocals, etc. (3) Echoes, the robot hears its own voice and is interrupted by itself (4) Misplaced-turn, the robot responds before the user has said the last word, which causes the robot to be interrupted by the misplaced utterance. To detect the above false barge-in, relying on audio or transcribed text alone is insufficient. Therefore, we build an end-to-end multimodal barge-in detection model to robustly detect whether the user has the intention to interrupt the robot.

When designing this feature, we consulted customer service professionals about whether robots should be allowed to interrupt users. Since most human customer services are not allowed to interrupt users, we did not design the function for robots to interrupt users when users are speaking.

Note that the inference timing for user state detection and backchannel selection is when the user is speaking, while barge-in detection is when the agent plays the audio stream. The former models perform inference on IPUs, while barge-in detection continually infers when intermediate results of streaming ASR change.  

\section{Modeling}
In this section, we describe the multimodal model used by user state detection and barge-in detection. The proposed method is shown in Figure \ref{fig:model}, and could be divided into four steps: feature extraction, multimodal data augmentation, modality fusion, and semi-supervised learning. 

\subsection{Feature Extraction}
Due to resource constraints, all modules are deployed on CPUs. Therefore, we choose lightweight models as feature encoders for lower CPU usage and faster runtime. 

\textbf{Text Encoder} First, the model inputs include user text $t_\text{user}$, user audio $a_\text{user}$, and bot previous response $t_\text{bot}$. We use a 1D convolutional neural network (CNN)\cite{kim-2014-convolutional}  with $k$ kernels followed by max-pooling to obtains representation of utterance $r_t$:
\begin{align}
    r_t = \text{MaxPooling}(\text{CNN}(t)) 
\end{align}
where $r_t \in \mathds R^{k*n_\text{filters}}$. We concatenate the extracted representation of the user and bot and feed into the fully-connected layer to obtain text features $f_t \in \mathds R^{H}$:
\begin{align}
    f_t = \text{ReLU}(W_1([r_\text{t, user}; r_\text{t,bot})])
\end{align}
where $ W_1 \in \mathds R^{H \times d}$ is learnable parameters, H is the hidden layer size, and d $\in \mathds R^{2k*n_\text{filters}}$ is the size of concatenating representation from user and bot.

\textbf{Audio Encoder} We extract audio features by using a single layer of gated recurrent unit (GRU) network. We take the last output vector of the GRU network as audio features $f_a \in \mathds R^{H}$, where H is the hidden layer size the same as text features.

We have tried using a single-layer transformer, LSTM, or bidirectional recurrent neural networks as the audio encoder. We found no significant differences between these models trained from scratch. Furthermore, the results of 1D-CNN on audio are slightly worse than the above models. Therefore, we eventually chose GRU as our audio encoder.

\subsection{Multimodal Data Augmentation}
In this section, we demonstrate how to perform multimodal data augmentation during training to improve the model generalization. 

First, we obtain sample $i$ from the data in the original order, and sample $j$ from the randomly shuffled order, where $i, j \in \{1, \dots, n \}$ and $n$ is the training batch size. Second, we mix the audio and text features of sample $i$ and $j$, respectively. We could obtain the augmented audio features $\hat f_a$ and text features $\hat f_t$:
\begin{align}
    \hat f_a &= f_{a,i}*\lambda + f_{a,j}*(1-\lambda) \\
    \hat f_t &= f_{t,i}*\lambda + f_{t,j}*(1-\lambda) 
\end{align}
where the $\lambda \in [0, 1]$ is the mixing ratio sampled from beta distribution: 
\begin{align}
    \lambda \sim Beta(\alpha, \alpha)
\end{align}
where $\alpha$ is the empirical hyper-parameter. Third, we also mix the corresponding label $y_i$ and $y_j$, and then get the augmented soft label  $\hat y$:
\begin{align}
     \hat y = y_i * \lambda + y_j * (1-\lambda)
\end{align}

Next, we will perform modality fusion and calculate the corresponding cross-entropy for classification.

\subsection{Modality Fusion}
Here we introduce the \textbf{bilinear gated fusion module} for modality fusion. First, we let text features $\hat f_t$ and audio features $\hat f_a$ interact through a gated linear unit, and only keep informative text features $\hat f_{t, g}$ and audio features $\hat f_{a, g}$:
\begin{align}
    \hat f_{t, g} &= \hat f_t \otimes \sigma(\hat f_a)  \\
    \hat f_{a, g} &= \hat f_a \otimes \sigma(\hat f_t)
\end{align}
where $\otimes$ is the dot product and $\sigma$ is the sigmoid function. 

Second, text features $\hat f_{t, g}$ and audio features $\hat f_{a, g}$ are fed to the bilinear layer to get multimodal features $f_m$: 
\begin{align}
    \hat f_m &= \hat f_{t, g} W_2 \hat f_{a, g} + b_2
\end{align}

Third, we feed multimodal features $\hat f_m$ into the classification layer with softmax activation to get the predicted probability $ p'$, and calculated the supervised cross-entropy $\mathcal{L}_\text{sup}$: 
\begin{align}
    \mathcal{L}_{\text{sup}} = - \sum_{k=1}^K \hat y_k \log \hat p_k - (1- \hat y_k) \log (1- \hat p_k)
\end{align}
where $K$ is the number of classes in the classification. 

\subsection{Semi-Supervised Learning}
We introduce semi-supervised learning methods to incorporate massive unlabeled data into multimodal models. First, we calculate the prediction result $p$ without data augmentation. Second, we calculate self-supervised label $\hat y_\text{semi}$ as follow:
\begin{align}
 y_\text{semi} :=\begin{cases}
            1, \quad \text{if} \quad p  > p_\text{threshold} \\
            0, \quad \text{else}
            \end{cases} 
\end{align}
where $p_\text{threshold}$ is an empirical probability threshold. If the values of $y_\text{semi}$ are all 0, the sample does not participate in the calculation of the loss function.

Finally, we calculate the cross-entropy for semi-supervised learning $\mathcal{L}_\text{semi}$ between prediction $\hat p$ with data augmentation and $y_\text{semi}$. Here we define auxiliary semi-supervised cross-entropy $\mathcal{L}_\text{semi}$ as follows: 
\vspace*{-0.5\baselineskip}
\begin{equation}
\begin{aligned}
    \mathcal{L}_{\text{semi}} = - \sum_{k=1}^K \hat y_{\text{semi}, k} \log \hat p_k - (1- \hat y_{\text{semi}, k}) \log (1- \hat p_k)
\end{aligned}
\end{equation}

For inference, we will not perform data augmentation, and use the features $f_t, f_a, f_m$ without MDA to obtain the predicted probability $p$, and calculate the classification result $y$ as follow:
\begin{align}
    y = \argmax_p p
\end{align}

\section{Experiments}
\subsection{Datasets}
Since there are no large-scale public multi-turn and multi-modal benchmark datasets, we construct three in-house datasets and conduct experiments. We extract the audio-text pairs for labeling from 100k crowdsourced Chinese conversation recordings with transcripts, including human-to-human and human-machine interactions. Note that each word in the transcript has a corresponding timestamp for alignment. All data is recorded, obtained, and used under the consent of users.

\subsubsection{User State Detection} It is a multi-class classification task containing three types of user states, including turn-switch, turn-keep, and turn-keep with hesitation. We labeled 30k data by crowdsourcing and introduced an additional 90k unlabeled data from in-house audio clips. Each clip contains 1 to 5 seconds of the latest user audio, current user transcript, and bot's previous response text.

\subsubsection{Backchannel Selection} We construct a multi-label classification task with 70k utterances containing ten kinds of common backchannel responses mining from corpus, such as \textit{Um-hum, Sure, Well} $\cdots$. Note that we only use text as input in this task.

\subsubsection{Barge-in Detection} A binary classification task to determine whether the user wants to interrupt the machine or not. We labeled 10k data by crowdsourcing and introduced an additional 10k noise clips from MUSAN dataset \cite{snyder2015musan} and 100k unlabeled data from in-house audio clips. Similarly, each clip contains 1 to 5 seconds of the latest user audio, current user transcript, and the bot's previous response text. 

The crowdsourcing annotation accuracy for user state detection and barge-in is around 90\%. We split the dataset into train, dev, and test sets in a ratio of 80:10:10 and report metrics on the test set. 

\subsection{Evaluation Metrics}
For evaluation metrics, we adopt metrics widely used in previous studies for classification tasks. For user state detection, we use accuracy and macro F1 score. For backchannel selection, we adopt  Hamming Loss \cite{tsoumakas2006multi} for multi-label classification. 

We also use manual labeled correctness, which could be viewed as accuracy, as the human evaluation metrics. For barge-in detection, we adopt precision, recall and macro F1 score. All metrics range from 0 to 1 and are displayed as percentages (\%). The higher the score, the better the performance, except that the Hamming loss is lower the better.

\begin{table}
  \caption{The experimental results of user state detection.}
  \label{tab:speaking_state}
    \begin{tabular}{llccc}
    \toprule
        \# & Model                 & Accuracy & F1   \\
    \midrule
        1          & Text encoder             & 76.39          & 62.31          \\
        2          & Audio encoder            & 79.47          & 58.14          \\
        3          & (1) + (2)                & 86.47          & 77.06          \\
        4          & (3) + Dialogue context     & 87.34          & 78.63          \\
        5          & (3) + MDA                & 88.32          & 79.92          \\
        6          & (3) + Semi-supervised    & 88.77          & 82.73          \\
        \textbf{7} & \textbf{(4) + (5) + (6)} & \textbf{91.05} & \textbf{85.08} \\
    \bottomrule
    \end{tabular}
\end{table}

\begin{table}
  \caption{The experimental results of barge-in detection.}
  \label{tab:bargein}
  \begin{tabular}{llccc}
    \toprule
    \# & Model                 & Precision & Recall & F1 \\
    \midrule
        1          & Text encoder                   & 53.96          & 42.79          & 47.73          \\
        2          & Audio encoder                  & 78.40          & 74.35          & 76.32          \\
        3          & (1) + (2)                      & 89.38          & 70.01          & 78.52          \\
        4          & (3) + Barge-in timing          & 90.02          & 73.97          & 81.21          \\
        5          & (3) + Dialogue context           & 89.95          & 76.60          & 82.74          \\
        6          & (3) + MDA                      & 89.91          & 78.70          & 83.93          \\
        7          & (3) + Semi-supervised          & 89.26          & 83.20          & 86.12          \\
        \textbf{8} & \textbf{(4) + (5) + (6) + (7)} & \textbf{91.27} & \textbf{86.21} & \textbf{88.67} \\
    \bottomrule
  \end{tabular}
\end{table}

\subsection{Results and Discussion}
We show the results of user state detection, barge-in, and backchannel selection in Table \ref{tab:speaking_state}, \ref{tab:bargein}, and \ref{tab:backchannel}, respectively. The tables show that the proposed method achieves the best results on all three tasks.

\subsubsection{Overall Results}
For user state detection, as shown in Table \ref{tab:speaking_state}, the accuracy of our proposed model reaches 91.05\% in accuracy, which is 4.58\% absolute improvement compared with the audio-text fusion baseline. Through qualitative analysis, we found that by joint modeling of audio and text, the model could more effectively capture subtle acoustic features, such as whether the user pitch gradually decreases, or hesitation, gasping, and stuttering.

\begin{table}
  \caption{The experimental results of backchannel selection.}
  \label{tab:backchannel}
    \begin{tabular}{cccc}
    \toprule
        \# & Method                              & Correctness   & $\mathcal{L}_\text{{Hamming}}$ \\
    \midrule
        1 & Weighted random                      & 77.6          & 23.74                          \\
        2 & Single-label                         & 84.3          & 21.94                          \\
        3 & Multi-label                          & 90.1          & 11.04                          \\
        4 & \textbf{Multi-label + w/ Soft-label} & \textbf{91.2} & \textbf{10.53}                 \\
    \bottomrule
    \end{tabular}
\end{table}

\begin{table*}
  \caption{Online A/B experiments on time consuming analysis with or without backchannel responses.}
  \label{tab:time_analysis}
    \begin{tabular}{ccc}
    \toprule
        Procedure                                        & w/o Backchannel & w/ Backchannel  \\
    \midrule
        Establish connection and transmit audio stream to ASR & 100ms      & 100ms           \\
        ASR Processing                                   & 150ms           & 150ms           \\
        VAD silence threshold                            & 800ms           & 200ms           \\
        VAD delay                                        & 100ms           & 100ms           \\
        Request Core Dialog Engine                       & 50ms            & -               \\
        Request Duplex Conversation                      & -               & 10ms            \\
        Request TTS                                      & 100ms           & 40ms           \\
        Return the audio stream and play                 & 100ms           & 100ms           \\
    \midrule
        \textbf{Total latency}                           & 1400ms & \textbf{700ms} \\
    \bottomrule
    \end{tabular}
\end{table*}

We show the results of barge-in detection in Table \ref{tab:bargein}. The proposed method achieves an absolute improvement of 16.2\% in recall and 10.15\% in F1 score compared with the baseline method while maintaining the precision around 90\%. Experimental results show that our multimodal model is able to identify false barge-in requests and is robust to various noises.

For the backchannel selection in Table \ref{tab:backchannel}, the proposed multi-label with soft-label method also achieved 91.2\% correctness, an absolute improvement of about 6.9\% compared with the single-label baseline. The agent's response latency could be significantly reduced by inserting the backchannel response before the answer. Table \ref{tab:time_analysis} is the latency analysis in the online environment where our system has been deployed. The results show that the response latency is reduced from 1400ms to 700ms, saving 50\% of user waiting time and resulting in a better user experience.

In the following discussion, we will focus on the multimodal models for in-depth analysis.

\subsubsection{Importance of Different Modality}
We investigate the influence of different modalities on different tasks. In user state detection, both audio and text modes are essential. Audio or text alone cannot achieve satisfactory results, and the accuracy is even lower than 80\%. In barge-in detection, the importance of audio is much greater than that text, and the F1 score of the text model is below 50\%. In spoken dialogue systems, it is difficult to determine whether the recognized speech is from a normal customer query or background noise based on the transcribed text alone.

To sum up, experiments show that relying solely on text or speech modalities cannot achieve satisfactory user state detection or barge-in detection results. The multimodal model can make good use of the information of different modalities and achieve better results. 

\subsubsection{Influence of Multimodal Data Augmentation (MDA)}
We further discuss the improvements brought about by multimodal data augmentation. In user state detection, MDA can bring an absolute accuracy improvement of 1.85\%. However, MDA can significantly improve the F1-score in barge-in detection, resulting in a 5.41 improvement compared to the multimodal baseline. 

We speculate there are two possible reasons why MDA achieves better improvement on barge-in detection. First, in barge-in detection, noise and non-noise classes are complimentary, and MDA can effectively improve the diversity of data, thereby making the results more robust. Second, by introducing additional external noise data in the barge-in detection, this part of the data plays an important role and improves the generalization ability of the model. 

\subsubsection{Influence of Semi-supervised Learning (SSL)}
Whether user state or barge-in detection, the performance is greatly improved after introducing massive unlabeled data through semi-supervised learning. Compared with the multimodal baseline, SSL achieves f1 score improvements of 5.67 and 7.6 in user state detection and barge-in detection, respectively. The experimental results show that in multimodal models, semi-supervised learning can make good use of massive unlabeled data, which has great potential to improve model performance.

\subsection{Deployment Lessons}
The proposed method has been deployed in Alibaba intelligent customer service for more than half a year, serving dozens of enterprise customers in different businesses. The following are our lessons learned during the deployment and application.

\subsubsection{Latency leads to inconsistent online and offline data distribution}
Due to the latency of streaming ASR recognition in an online environment, there is a delay between transcribed text and audio, resulting in data misalignment. There is typically a 300 to 600 ms delay between audio and text in our deployment experiments, and audio is faster than text. This delay will lead to the inconsistent distribution of offline data and online traffic, which will lead to poor model results. 

We take two empirical approaches to solve it. The first method is to reduce ASR latency as much as possible, such as deploying middleware and models on the same machine. The second method is to simulate the misalignment of online text and audio in advance when constructing offline training data to ensure that the distribution of offline data and online traffic is consistent. 

Another way is to align older audio streams with text intentionally. However, we found that using the latest audio stream for inference - even without text alignment, resulted in better model performance, so we did not go with the third method.

\subsubsection{System throughput}
We observe that the number of ASR requests will double when the duplex conversation capability is activated. Most of the pressure comes from streaming recognition for barge-in detection. The maximum system throughput will be reduced to half of the original.

\subsubsection{Middleware support}
The duplex conversation cannot be deployed if the online environment only adopts the traditional MRCP (Media Resource Control Protocol) without customized middleware. We rely on customized Interactive Voice Response (IVR) middleware to send customized audio streams and transcribed text to our models. 

\section{Related Work} 
Turn-taking \cite{raux2008flexible} is the core concept of duplex conversation. It has been extensively studied in different fields, including linguistics, phonetics, and sociology for the past decades \cite{skantze2021turn}. The user state detection we proposed could be considered as a variant of turn-taking behavior. Previous studies \cite{raux2009finite, sacks1978simplest} use a non-deterministic finite-state machine with six states to describe the turn-taking behavior between system and user in spoken dialogue systems (SDS). It illustrates all possible states of turn-taking in SDS, and defines the goal of turn-taking is to minimize the time of mutual silence or speech between the two interlocutors, thereby improving communication efficiency. 

There are three essential concepts of turn-taking. The first is turn-taking cues \cite{duncan1972some, duncan1974signalling}, including speech, rhythm, breathing, gaze, or gesture. The agents can use these turn-taking cues to determine whether to take the turn from the user, or the agent can use cues to release the turn. The second is end-of-turn detection \cite{lala2017attentive, hara2019turn, chen2021human} or prediction \cite{ekstedt-skantze-2020-turngpt, lala2019smooth, razavi2019investigating}. The difference between detection and prediction is that detection determines whether the agent should take a turn at the present moment. In contrast, prediction determines when to take a turn in the future. Note that our proposed user state detection falls into the former category. The third is overlap, which mainly includes two situations. When the speech of user and agent overlap, if the user wants to take the turn from agents, then we define the behavior as an interruption, or barge-in \cite{zhao2015incremental, selfridge2013continuously, matsuyama2009enabling, khouzaimi2016reinforcement}. If the user has no intention to take the turn, we call the behavior backchannel or listener responses \cite{hara2018prediction, yngve1970getting}, such as \textit{"Um-hum, Yeah, Right"}. We could have a deeper understanding of turn-taking behavior in duplex conversation through the above concepts.

Traditional dialogue systems \cite{liu2021dialoguecse, he2022unified} usually consist of three components: natural language understanding (NLU) \cite{lin2020discovering, Zhang_Xu_Lin_Lyu_2021, lin2019deep, Zhang_Xu_Lin_2021}, dialogue management (DM) \cite{dai2020learning, dai2021preview, he2021galaxy}, and natural language generation (NLG) \cite{zhou2021eva, wang2021diversifying, zheng2020pre, zhao-etal-2022-improving} modules. Empirically, NLU plays the most important role in task-oriented dialogue systems, including tasks such as intent detection \cite{geng2019induction, geng2020dynamic, lin2019post, zhang-etal-2021-textoir}, slot filling \cite{zhang-etal-2022-slot}, and semantic parsing \cite{hui2021dynamic, hui-etal-2022-s2sql, wang2022proton}. In spoken dialogue system , spoken language understanding (SLU) can be viewed as a subset of NLU. Most studies \cite{faruqui2021revisiting, marge2022spoken} ignore audio modality and focus only on transcribed text obtained through ASR and treat it as an NLU task. In this work, we leverage speech and transcribed text to jointly capture complex behaviors beyond words in human-machine interaction for duplex conversation. 

Multimodal modeling is a research hotspot that has attracted the attention of many scholars in recent years \cite{baltruvsaitis2018multimodal, zheng2021mmchat}. The key to multimodal modeling includes multimodal fusion \cite{zadeh-etal-2017-tensor, YangWYZRZPM21},  consistency and difference \cite{HazarikaZP20, yu2021learning}, modality alignment \cite{TsaiBLKMS19}. We recommend the survey \cite{baltruvsaitis2018multimodal} for a comprehensive understanding. Semi-supervised learning (SSL) \cite{andrade2021survey} is also an area that has attracted much attention in the field of machine learning in recent years. Modern semi-supervised learning methods in deep learning are based on consistency regularization and entropy minimization. Most of the method utilize data augmentation \cite{zhang2020dialogue} to create learning objective for consistency regularization, such as MixMatch \cite{berthelot2019mixmatch}, UDA \cite{XieDHL020}, ReMixMatch \cite{berthelot2019remixmatch}, and FixMatch \cite{SohnBCZZRCKL20}. Our approach to SSL is closest to FixMatch in computer vision. We extend the FixMatch method from images to the multimodal scenario, using different data augmentation methods and loss functions.

\section{Conclusion}
In this paper, we present \textit{Duplex Conversation}, a telephone-based multi-turn, multimodal spoken dialogue system that enables agents to communicate with customers in human-like behavior. We demonstrate what a full-duplex conversation should look like and how we build full-duplex capabilities through three subtasks. Furthermore, we propose a multimodal data augmentation method that effectively improves model robustness and domain generalization by leveraging massive unlabeled data through semi-supervised learning. Experimental results on three in-house datasets show that the proposed method outperforms multimodal baselines by a large margin. Online A/B experiments show that our duplex conversation could significantly reduce response latency by 50\%.

In the future, two promising directions are worth noting for duplex conversation. One is introducing reinforcement learning to update parameters online. Another is the application beyond telephone agents, such as digital human agents with multimodal interactions, including real-time audio-visual responses such as gaze, facial expressions, and body language, for smooth turn-taking and human-like experiences. 

\begin{acks}
Duplex Conversation has contributions from members of Alibaba intelligent customer service team, notably Kai Cheng, Yifan Xie, Ke Yan, Jia Tan, Jin Zhu, Xiangfeng Cheng, and Can Li. We would like to thank Yinpei Dai and anonymous reviewers for the constructive comments.
\end{acks}

\bibliographystyle{ACM-Reference-Format}
\balance
\bibliography{sample-base}


\begin{thebibliography}{67}


\ifx \showCODEN    \undefined \def \showCODEN     #1{\unskip}     \fi
\ifx \showDOI      \undefined \def \showDOI       #1{#1}\fi
\ifx \showISBNx    \undefined \def \showISBNx     #1{\unskip}     \fi
\ifx \showISBNxiii \undefined \def \showISBNxiii  #1{\unskip}     \fi
\ifx \showISSN     \undefined \def \showISSN      #1{\unskip}     \fi
\ifx \showLCCN     \undefined \def \showLCCN      #1{\unskip}     \fi
\ifx \shownote     \undefined \def \shownote      #1{#1}          \fi
\ifx \showarticletitle \undefined \def \showarticletitle #1{#1}   \fi
\ifx \showURL      \undefined \def \showURL       {\relax}        \fi
\providecommand\bibfield[2]{#2}
\providecommand\bibinfo[2]{#2}
\providecommand\natexlab[1]{#1}
\providecommand\showeprint[2][]{arXiv:#2}

\bibitem[\protect\citeauthoryear{Andrade, Rodrigues, and Novais}{Andrade
  et~al\mbox{.}}{2021}]%
        {andrade2021survey}
\bibfield{author}{\bibinfo{person}{Guilherme Andrade}, \bibinfo{person}{Manuel
  Rodrigues}, {and} \bibinfo{person}{Paulo Novais}.}
  \bibinfo{year}{2021}\natexlab{}.
\newblock \showarticletitle{A Survey on the Semi Supervised Learning Paradigm
  in the Context of Speech Emotion Recognition}. In
  \bibinfo{booktitle}{\emph{Proceedings of SAI Intelligent Systems
  Conference}}. Springer, \bibinfo{pages}{771--792}.
\newblock


\bibitem[\protect\citeauthoryear{Baltru{\v{s}}aitis, Ahuja, and
  Morency}{Baltru{\v{s}}aitis et~al\mbox{.}}{2018}]%
        {baltruvsaitis2018multimodal}
\bibfield{author}{\bibinfo{person}{Tadas Baltru{\v{s}}aitis},
  \bibinfo{person}{Chaitanya Ahuja}, {and} \bibinfo{person}{Louis-Philippe
  Morency}.} \bibinfo{year}{2018}\natexlab{}.
\newblock \showarticletitle{Multimodal machine learning: A survey and
  taxonomy}.
\newblock \bibinfo{journal}{\emph{IEEE transactions on pattern analysis and
  machine intelligence}} \bibinfo{volume}{41}, \bibinfo{number}{2}
  (\bibinfo{year}{2018}), \bibinfo{pages}{423--443}.
\newblock


\bibitem[\protect\citeauthoryear{Berthelot, Carlini, Cubuk, Kurakin, Sohn,
  Zhang, and Raffel}{Berthelot et~al\mbox{.}}{2019a}]%
        {berthelot2019remixmatch}
\bibfield{author}{\bibinfo{person}{David Berthelot}, \bibinfo{person}{Nicholas
  Carlini}, \bibinfo{person}{Ekin~D Cubuk}, \bibinfo{person}{Alex Kurakin},
  \bibinfo{person}{Kihyuk Sohn}, \bibinfo{person}{Han Zhang}, {and}
  \bibinfo{person}{Colin Raffel}.} \bibinfo{year}{2019}\natexlab{a}.
\newblock \showarticletitle{Remixmatch: Semi-supervised learning with
  distribution alignment and augmentation anchoring}.
\newblock \bibinfo{journal}{\emph{arXiv preprint arXiv:1911.09785}}
  (\bibinfo{year}{2019}).
\newblock


\bibitem[\protect\citeauthoryear{Berthelot, Carlini, Goodfellow, Papernot,
  Oliver, and Raffel}{Berthelot et~al\mbox{.}}{2019b}]%
        {berthelot2019mixmatch}
\bibfield{author}{\bibinfo{person}{David Berthelot}, \bibinfo{person}{Nicholas
  Carlini}, \bibinfo{person}{Ian Goodfellow}, \bibinfo{person}{Nicolas
  Papernot}, \bibinfo{person}{Avital Oliver}, {and} \bibinfo{person}{Colin~A
  Raffel}.} \bibinfo{year}{2019}\natexlab{b}.
\newblock \showarticletitle{Mixmatch: A holistic approach to semi-supervised
  learning}.
\newblock \bibinfo{journal}{\emph{Advances in Neural Information Processing
  Systems}}  \bibinfo{volume}{32} (\bibinfo{year}{2019}).
\newblock


\bibitem[\protect\citeauthoryear{Chen, Li, Dai, Zhou, and Chen}{Chen
  et~al\mbox{.}}{2021}]%
        {chen2021human}
\bibfield{author}{\bibinfo{person}{Kehan Chen}, \bibinfo{person}{Zezhong Li},
  \bibinfo{person}{Suyang Dai}, \bibinfo{person}{Wei Zhou}, {and}
  \bibinfo{person}{Haiqing Chen}.} \bibinfo{year}{2021}\natexlab{}.
\newblock \showarticletitle{Human-to-Human Conversation Dataset for Learning
  Fine-grained Turn-taking Action}.
\newblock \bibinfo{journal}{\emph{Proc. Interspeech 2021}}
  (\bibinfo{year}{2021}), \bibinfo{pages}{3231--3235}.
\newblock


\bibitem[\protect\citeauthoryear{Dai, Li, Li, Sun, Huang, Si, and Zhu}{Dai
  et~al\mbox{.}}{2021}]%
        {dai2021preview}
\bibfield{author}{\bibinfo{person}{Yinpei Dai}, \bibinfo{person}{Hangyu Li},
  \bibinfo{person}{Yongbin Li}, \bibinfo{person}{Jian Sun},
  \bibinfo{person}{Fei Huang}, \bibinfo{person}{Luo Si}, {and}
  \bibinfo{person}{Xiaodan Zhu}.} \bibinfo{year}{2021}\natexlab{}.
\newblock \showarticletitle{Preview, Attend and Review: Schema-Aware Curriculum
  Learning for Multi-Domain Dialogue State Tracking}. In
  \bibinfo{booktitle}{\emph{Proceedings of the 59th Annual Meeting of the
  Association for Computational Linguistics and the 11th International Joint
  Conference on Natural Language Processing (Volume 2: Short Papers)}}.
  \bibinfo{pages}{879--885}.
\newblock


\bibitem[\protect\citeauthoryear{Dai, Li, Tang, Li, Sun, and Zhu}{Dai
  et~al\mbox{.}}{2020}]%
        {dai2020learning}
\bibfield{author}{\bibinfo{person}{Yinpei Dai}, \bibinfo{person}{Hangyu Li},
  \bibinfo{person}{Chengguang Tang}, \bibinfo{person}{Yongbin Li},
  \bibinfo{person}{Jian Sun}, {and} \bibinfo{person}{Xiaodan Zhu}.}
  \bibinfo{year}{2020}\natexlab{}.
\newblock \showarticletitle{Learning low-resource end-to-end goal-oriented
  dialog for fast and reliable system deployment}. In
  \bibinfo{booktitle}{\emph{Proceedings of the 58th Annual Meeting of the
  Association for Computational Linguistics}}. \bibinfo{pages}{609--618}.
\newblock


\bibitem[\protect\citeauthoryear{Duncan}{Duncan}{1972}]%
        {duncan1972some}
\bibfield{author}{\bibinfo{person}{Starkey Duncan}.}
  \bibinfo{year}{1972}\natexlab{}.
\newblock \showarticletitle{Some signals and rules for taking speaking turns in
  conversations.}
\newblock \bibinfo{journal}{\emph{Journal of personality and social
  psychology}} \bibinfo{volume}{23}, \bibinfo{number}{2}
  (\bibinfo{year}{1972}), \bibinfo{pages}{283}.
\newblock


\bibitem[\protect\citeauthoryear{Duncan~Jr and Niederehe}{Duncan~Jr and
  Niederehe}{1974}]%
        {duncan1974signalling}
\bibfield{author}{\bibinfo{person}{Starkey Duncan~Jr} {and}
  \bibinfo{person}{George Niederehe}.} \bibinfo{year}{1974}\natexlab{}.
\newblock \showarticletitle{On signalling that it's your turn to speak}.
\newblock \bibinfo{journal}{\emph{Journal of experimental social psychology}}
  \bibinfo{volume}{10}, \bibinfo{number}{3} (\bibinfo{year}{1974}),
  \bibinfo{pages}{234--247}.
\newblock


\bibitem[\protect\citeauthoryear{Ekstedt and Skantze}{Ekstedt and
  Skantze}{2020}]%
        {ekstedt-skantze-2020-turngpt}
\bibfield{author}{\bibinfo{person}{Erik Ekstedt} {and} \bibinfo{person}{Gabriel
  Skantze}.} \bibinfo{year}{2020}\natexlab{}.
\newblock \showarticletitle{{T}urn{GPT}: a Transformer-based Language Model for
  Predicting Turn-taking in Spoken Dialog}. In
  \bibinfo{booktitle}{\emph{Findings of the Association for Computational
  Linguistics: EMNLP 2020}}. \bibinfo{publisher}{Association for Computational
  Linguistics}, \bibinfo{pages}{2981--2990}.
\newblock


\bibitem[\protect\citeauthoryear{Faruqui and Hakkani-T{\"u}r}{Faruqui and
  Hakkani-T{\"u}r}{2021}]%
        {faruqui2021revisiting}
\bibfield{author}{\bibinfo{person}{Manaal Faruqui} {and} \bibinfo{person}{Dilek
  Hakkani-T{\"u}r}.} \bibinfo{year}{2021}\natexlab{}.
\newblock \showarticletitle{Revisiting the Boundary between ASR and NLU in the
  Age of Conversational Dialog Systems}.
\newblock \bibinfo{journal}{\emph{Computational Linguistics}}
  (\bibinfo{year}{2021}), \bibinfo{pages}{1--12}.
\newblock


\bibitem[\protect\citeauthoryear{Geng, Li, Li, Sun, and Zhu}{Geng
  et~al\mbox{.}}{2020}]%
        {geng2020dynamic}
\bibfield{author}{\bibinfo{person}{Ruiying Geng}, \bibinfo{person}{Binhua Li},
  \bibinfo{person}{Yongbin Li}, \bibinfo{person}{Jian Sun}, {and}
  \bibinfo{person}{Xiaodan Zhu}.} \bibinfo{year}{2020}\natexlab{}.
\newblock \showarticletitle{Dynamic Memory Induction Networks for Few-Shot Text
  Classification}. In \bibinfo{booktitle}{\emph{Proceedings of the 58th Annual
  Meeting of the Association for Computational Linguistics}}.
  \bibinfo{pages}{1087--1094}.
\newblock


\bibitem[\protect\citeauthoryear{Geng, Li, Li, Zhu, Jian, and Sun}{Geng
  et~al\mbox{.}}{2019}]%
        {geng2019induction}
\bibfield{author}{\bibinfo{person}{Ruiying Geng}, \bibinfo{person}{Binhua Li},
  \bibinfo{person}{Yongbin Li}, \bibinfo{person}{Xiaodan Zhu},
  \bibinfo{person}{Ping Jian}, {and} \bibinfo{person}{Jian Sun}.}
  \bibinfo{year}{2019}\natexlab{}.
\newblock \showarticletitle{Induction Networks for Few-Shot Text
  Classification}. In \bibinfo{booktitle}{\emph{Proceedings of the 2019
  Conference on Empirical Methods in Natural Language Processing and the 9th
  International Joint Conference on Natural Language Processing
  (EMNLP-IJCNLP)}}. \bibinfo{pages}{3904--3913}.
\newblock


\bibitem[\protect\citeauthoryear{Hara, Inoue, Takanashi, and Kawahara}{Hara
  et~al\mbox{.}}{2018}]%
        {hara2018prediction}
\bibfield{author}{\bibinfo{person}{Kohei Hara}, \bibinfo{person}{Koji Inoue},
  \bibinfo{person}{Katsuya Takanashi}, {and} \bibinfo{person}{Tatsuya
  Kawahara}.} \bibinfo{year}{2018}\natexlab{}.
\newblock \showarticletitle{Prediction of turn-taking using multitask learning
  with prediction of backchannels and fillers}.
\newblock \bibinfo{journal}{\emph{Listener}}  \bibinfo{volume}{162}
  (\bibinfo{year}{2018}), \bibinfo{pages}{364}.
\newblock


\bibitem[\protect\citeauthoryear{Hara, Inoue, Takanashi, and Kawahara}{Hara
  et~al\mbox{.}}{2019}]%
        {hara2019turn}
\bibfield{author}{\bibinfo{person}{Kohei Hara}, \bibinfo{person}{Koji Inoue},
  \bibinfo{person}{Katsuya Takanashi}, {and} \bibinfo{person}{Tatsuya
  Kawahara}.} \bibinfo{year}{2019}\natexlab{}.
\newblock \showarticletitle{Turn-Taking Prediction Based on Detection of
  Transition Relevance Place.}. In \bibinfo{booktitle}{\emph{INTERSPEECH}}.
  \bibinfo{pages}{4170--4174}.
\newblock


\bibitem[\protect\citeauthoryear{Hazarika, Zimmermann, and Poria}{Hazarika
  et~al\mbox{.}}{2020}]%
        {HazarikaZP20}
\bibfield{author}{\bibinfo{person}{Devamanyu Hazarika}, \bibinfo{person}{Roger
  Zimmermann}, {and} \bibinfo{person}{Soujanya Poria}.}
  \bibinfo{year}{2020}\natexlab{}.
\newblock \showarticletitle{Misa: Modality-invariant and-specific
  representations for multimodal sentiment analysis}. In
  \bibinfo{booktitle}{\emph{Proceedings of the 28th ACM International
  Conference on Multimedia}}. \bibinfo{pages}{1122--1131}.
\newblock


\bibitem[\protect\citeauthoryear{He, Dai, Yang, Huang, Si, Sun, and Li}{He
  et~al\mbox{.}}{2022}]%
        {he2022unified}
\bibfield{author}{\bibinfo{person}{Wanwei He}, \bibinfo{person}{Yinpei Dai},
  \bibinfo{person}{Min Yang}, \bibinfo{person}{Fei Huang}, \bibinfo{person}{Luo
  Si}, \bibinfo{person}{jian Sun}, {and} \bibinfo{person}{Yongbin Li}.}
  \bibinfo{year}{2022}\natexlab{}.
\newblock \showarticletitle{Unified Dialog Model Pre-training for Task-Oriented
  Dialog Understanding and Generation}.
\newblock \bibinfo{journal}{\emph{SIGIR}} (\bibinfo{year}{2022}).
\newblock


\bibitem[\protect\citeauthoryear{He, Dai, Zheng, Wu, Cao, Liu, Jiang, Yang,
  Huang, Si, et~al\mbox{.}}{He et~al\mbox{.}}{2021}]%
        {he2021galaxy}
\bibfield{author}{\bibinfo{person}{Wanwei He}, \bibinfo{person}{Yinpei Dai},
  \bibinfo{person}{Yinhe Zheng}, \bibinfo{person}{Yuchuan Wu},
  \bibinfo{person}{Zheng Cao}, \bibinfo{person}{Dermot Liu},
  \bibinfo{person}{Peng Jiang}, \bibinfo{person}{Min Yang},
  \bibinfo{person}{Fei Huang}, \bibinfo{person}{Luo Si}, {et~al\mbox{.}}}
  \bibinfo{year}{2021}\natexlab{}.
\newblock \showarticletitle{GALAXY: A Generative Pre-trained Model for
  Task-Oriented Dialog with Semi-Supervised Learning and Explicit Policy
  Injection}.
\newblock \bibinfo{journal}{\emph{arXiv preprint arXiv:2111.14592}}
  (\bibinfo{year}{2021}).
\newblock


\bibitem[\protect\citeauthoryear{Hui, Geng, Ren, Li, Li, Sun, Huang, Si, Zhu,
  and Zhu}{Hui et~al\mbox{.}}{2021}]%
        {hui2021dynamic}
\bibfield{author}{\bibinfo{person}{Binyuan Hui}, \bibinfo{person}{Ruiying
  Geng}, \bibinfo{person}{Qiyu Ren}, \bibinfo{person}{Binhua Li},
  \bibinfo{person}{Yongbin Li}, \bibinfo{person}{Jian Sun},
  \bibinfo{person}{Fei Huang}, \bibinfo{person}{Luo Si},
  \bibinfo{person}{Pengfei Zhu}, {and} \bibinfo{person}{Xiaodan Zhu}.}
  \bibinfo{year}{2021}\natexlab{}.
\newblock \showarticletitle{Dynamic hybrid relation exploration network for
  cross-domain context-dependent semantic parsing}. In
  \bibinfo{booktitle}{\emph{Proceedings of the AAAI Conference on Artificial
  Intelligence}}, Vol.~\bibinfo{volume}{35}. \bibinfo{pages}{13116--13124}.
\newblock


\bibitem[\protect\citeauthoryear{Hui, Geng, Wang, Qin, Li, Li, Sun, and Li}{Hui
  et~al\mbox{.}}{2022}]%
        {hui-etal-2022-s2sql}
\bibfield{author}{\bibinfo{person}{Binyuan Hui}, \bibinfo{person}{Ruiying
  Geng}, \bibinfo{person}{Lihan Wang}, \bibinfo{person}{Bowen Qin},
  \bibinfo{person}{Yanyang Li}, \bibinfo{person}{Bowen Li},
  \bibinfo{person}{Jian Sun}, {and} \bibinfo{person}{Yongbin Li}.}
  \bibinfo{year}{2022}\natexlab{}.
\newblock \showarticletitle{{S}$^2${SQL}: Injecting Syntax to Question-Schema
  Interaction Graph Encoder for Text-to-{SQL} Parsers}. In
  \bibinfo{booktitle}{\emph{Findings of the Association for Computational
  Linguistics: ACL 2022}}. \bibinfo{publisher}{Association for Computational
  Linguistics}, \bibinfo{address}{Dublin, Ireland},
  \bibinfo{pages}{1254--1262}.
\newblock


\bibitem[\protect\citeauthoryear{Inoue, Lala, Yamamoto, Nakamura, Takanashi,
  and Kawahara}{Inoue et~al\mbox{.}}{2020}]%
        {inoue2020attentive}
\bibfield{author}{\bibinfo{person}{Koji Inoue}, \bibinfo{person}{Divesh Lala},
  \bibinfo{person}{Kenta Yamamoto}, \bibinfo{person}{Shizuka Nakamura},
  \bibinfo{person}{Katsuya Takanashi}, {and} \bibinfo{person}{Tatsuya
  Kawahara}.} \bibinfo{year}{2020}\natexlab{}.
\newblock \showarticletitle{An attentive listening system with android ERICA:
  Comparison of autonomous and WOZ interactions}. In
  \bibinfo{booktitle}{\emph{Proceedings of the 21th Annual Meeting of the
  Special Interest Group on Discourse and Dialogue}}.
  \bibinfo{pages}{118--127}.
\newblock


\bibitem[\protect\citeauthoryear{Jin, Yang, and Wen}{Jin et~al\mbox{.}}{2021}]%
        {jin2021duplex}
\bibfield{author}{\bibinfo{person}{Chunxiang Jin}, \bibinfo{person}{Minghui
  Yang}, {and} \bibinfo{person}{Zujie Wen}.} \bibinfo{year}{2021}\natexlab{}.
\newblock \showarticletitle{Duplex Conversation in Outbound Agent System}.
\newblock \bibinfo{journal}{\emph{Proc. Interspeech 2021}}
  (\bibinfo{year}{2021}), \bibinfo{pages}{4866--4867}.
\newblock


\bibitem[\protect\citeauthoryear{Khouzaimi, Laroche, and Lef{\`e}vre}{Khouzaimi
  et~al\mbox{.}}{2016}]%
        {khouzaimi2016reinforcement}
\bibfield{author}{\bibinfo{person}{Hatim Khouzaimi}, \bibinfo{person}{Romain
  Laroche}, {and} \bibinfo{person}{Fabrice Lef{\`e}vre}.}
  \bibinfo{year}{2016}\natexlab{}.
\newblock \showarticletitle{Reinforcement Learning for Turn-Taking Management
  in Incremental Spoken Dialogue Systems.}. In
  \bibinfo{booktitle}{\emph{IJCAI}}. \bibinfo{pages}{2831--2837}.
\newblock


\bibitem[\protect\citeauthoryear{Kim}{Kim}{2014}]%
        {kim-2014-convolutional}
\bibfield{author}{\bibinfo{person}{Yoon Kim}.} \bibinfo{year}{2014}\natexlab{}.
\newblock \showarticletitle{Convolutional Neural Networks for Sentence
  Classification}. In \bibinfo{booktitle}{\emph{Proceedings of the 2014
  Conference on Empirical Methods in Natural Language Processing ({EMNLP})}}.
  \bibinfo{publisher}{Association for Computational Linguistics},
  \bibinfo{address}{Doha, Qatar}, \bibinfo{pages}{1746--1751}.
\newblock


\bibitem[\protect\citeauthoryear{Lala, Inoue, and Kawahara}{Lala
  et~al\mbox{.}}{2019}]%
        {lala2019smooth}
\bibfield{author}{\bibinfo{person}{Divesh Lala}, \bibinfo{person}{Koji Inoue},
  {and} \bibinfo{person}{Tatsuya Kawahara}.} \bibinfo{year}{2019}\natexlab{}.
\newblock \showarticletitle{Smooth turn-taking by a robot using an online
  continuous model to generate turn-taking cues}. In
  \bibinfo{booktitle}{\emph{2019 International Conference on Multimodal
  Interaction}}. \bibinfo{pages}{226--234}.
\newblock


\bibitem[\protect\citeauthoryear{Lala, Milhorat, Inoue, Ishida, Takanashi, and
  Kawahara}{Lala et~al\mbox{.}}{2017}]%
        {lala2017attentive}
\bibfield{author}{\bibinfo{person}{Divesh Lala}, \bibinfo{person}{Pierrick
  Milhorat}, \bibinfo{person}{Koji Inoue}, \bibinfo{person}{Masanari Ishida},
  \bibinfo{person}{Katsuya Takanashi}, {and} \bibinfo{person}{Tatsuya
  Kawahara}.} \bibinfo{year}{2017}\natexlab{}.
\newblock \showarticletitle{Attentive listening system with backchanneling,
  response generation and flexible turn-taking}. In
  \bibinfo{booktitle}{\emph{Proceedings of the 18th Annual SIGdial Meeting on
  Discourse and Dialogue}}. \bibinfo{pages}{127--136}.
\newblock


\bibitem[\protect\citeauthoryear{Leviathan and Matias}{Leviathan and
  Matias}{2018}]%
        {leviathan2018google}
\bibfield{author}{\bibinfo{person}{Yaniv Leviathan} {and}
  \bibinfo{person}{Yossi Matias}.} \bibinfo{year}{2018}\natexlab{}.
\newblock \showarticletitle{Google Duplex: an AI system for accomplishing
  real-world tasks over the phone}.
\newblock  (\bibinfo{year}{2018}).
\newblock


\bibitem[\protect\citeauthoryear{Lin and Xu}{Lin and Xu}{2019a}]%
        {lin2019deep}
\bibfield{author}{\bibinfo{person}{Ting-En Lin} {and} \bibinfo{person}{Hua
  Xu}.} \bibinfo{year}{2019}\natexlab{a}.
\newblock \showarticletitle{Deep Unknown Intent Detection with Margin Loss}. In
  \bibinfo{booktitle}{\emph{Proceedings of the 57th Annual Meeting of the
  Association for Computational Linguistics}}. \bibinfo{pages}{5491--5496}.
\newblock


\bibitem[\protect\citeauthoryear{Lin and Xu}{Lin and Xu}{2019b}]%
        {lin2019post}
\bibfield{author}{\bibinfo{person}{Ting-En Lin} {and} \bibinfo{person}{Hua
  Xu}.} \bibinfo{year}{2019}\natexlab{b}.
\newblock \showarticletitle{A post-processing method for detecting unknown
  intent of dialogue system via pre-trained deep neural network classifier}.
\newblock \bibinfo{journal}{\emph{Knowledge-Based Systems}}
  \bibinfo{volume}{186} (\bibinfo{year}{2019}), \bibinfo{pages}{104979}.
\newblock


\bibitem[\protect\citeauthoryear{Lin, Xu, and Zhang}{Lin et~al\mbox{.}}{2020}]%
        {lin2020discovering}
\bibfield{author}{\bibinfo{person}{Ting-En Lin}, \bibinfo{person}{Hua Xu},
  {and} \bibinfo{person}{Hanlei Zhang}.} \bibinfo{year}{2020}\natexlab{}.
\newblock \showarticletitle{Discovering New Intents via Constrained Deep
  Adaptive Clustering with Cluster Refinement}. In
  \bibinfo{booktitle}{\emph{Proceedings of AAAI}}. \bibinfo{pages}{8360--8367}.
\newblock


\bibitem[\protect\citeauthoryear{Liu, Jiang, Xiong, Yang, and Ye}{Liu
  et~al\mbox{.}}{2020}]%
        {liu2020towards}
\bibfield{author}{\bibinfo{person}{Che Liu}, \bibinfo{person}{Junfeng Jiang},
  \bibinfo{person}{Chao Xiong}, \bibinfo{person}{Yi Yang}, {and}
  \bibinfo{person}{Jieping Ye}.} \bibinfo{year}{2020}\natexlab{}.
\newblock \showarticletitle{Towards building an intelligent chatbot for
  customer service: Learning to respond at the appropriate time}. In
  \bibinfo{booktitle}{\emph{Proceedings of the 26th ACM SIGKDD international
  conference on Knowledge Discovery \& Data Mining}}.
  \bibinfo{pages}{3377--3385}.
\newblock


\bibitem[\protect\citeauthoryear{Liu, Wang, Xu, Li, and Ye}{Liu
  et~al\mbox{.}}{2019}]%
        {liu2019automatic}
\bibfield{author}{\bibinfo{person}{Chunyi Liu}, \bibinfo{person}{Peng Wang},
  \bibinfo{person}{Jiang Xu}, \bibinfo{person}{Zang Li}, {and}
  \bibinfo{person}{Jieping Ye}.} \bibinfo{year}{2019}\natexlab{}.
\newblock \showarticletitle{Automatic dialogue summary generation for customer
  service}. In \bibinfo{booktitle}{\emph{Proceedings of the 25th ACM SIGKDD
  International Conference on Knowledge Discovery \& Data Mining}}.
  \bibinfo{pages}{1957--1965}.
\newblock


\bibitem[\protect\citeauthoryear{Liu, Wang, Liu, Sun, Huang, and Si}{Liu
  et~al\mbox{.}}{2021}]%
        {liu2021dialoguecse}
\bibfield{author}{\bibinfo{person}{Che Liu}, \bibinfo{person}{Rui Wang},
  \bibinfo{person}{Jinghua Liu}, \bibinfo{person}{Jian Sun},
  \bibinfo{person}{Fei Huang}, {and} \bibinfo{person}{Luo Si}.}
  \bibinfo{year}{2021}\natexlab{}.
\newblock \showarticletitle{DialogueCSE: Dialogue-based Contrastive Learning of
  Sentence Embeddings}. In \bibinfo{booktitle}{\emph{Proceedings of the 2021
  Conference on Empirical Methods in Natural Language Processing}}.
  \bibinfo{pages}{2396--2406}.
\newblock


\bibitem[\protect\citeauthoryear{Marge, Espy-Wilson, Ward, Alwan, Artzi,
  Bansal, Blankenship, Chai, Daum{\'e}~III, Dey, et~al\mbox{.}}{Marge
  et~al\mbox{.}}{2022}]%
        {marge2022spoken}
\bibfield{author}{\bibinfo{person}{Matthew Marge}, \bibinfo{person}{Carol
  Espy-Wilson}, \bibinfo{person}{Nigel~G Ward}, \bibinfo{person}{Abeer Alwan},
  \bibinfo{person}{Yoav Artzi}, \bibinfo{person}{Mohit Bansal},
  \bibinfo{person}{Gil Blankenship}, \bibinfo{person}{Joyce Chai},
  \bibinfo{person}{Hal Daum{\'e}~III}, \bibinfo{person}{Debadeepta Dey},
  {et~al\mbox{.}}} \bibinfo{year}{2022}\natexlab{}.
\newblock \showarticletitle{Spoken language interaction with robots:
  Recommendations for future research}.
\newblock \bibinfo{journal}{\emph{Computer Speech \& Language}}
  \bibinfo{volume}{71} (\bibinfo{year}{2022}), \bibinfo{pages}{101255}.
\newblock


\bibitem[\protect\citeauthoryear{Matsuyama, Komatani, Ogata, and
  Okuno}{Matsuyama et~al\mbox{.}}{2009}]%
        {matsuyama2009enabling}
\bibfield{author}{\bibinfo{person}{Kyoko Matsuyama}, \bibinfo{person}{Kazunori
  Komatani}, \bibinfo{person}{Tetsuya Ogata}, {and} \bibinfo{person}{Hiroshi~G
  Okuno}.} \bibinfo{year}{2009}\natexlab{}.
\newblock \showarticletitle{Enabling a user to specify an item at any time
  during system enumeration-item identification for barge-in-able
  conversational dialogue systems}. In \bibinfo{booktitle}{\emph{Tenth Annual
  Conference of the International Speech Communication Association}}.
\newblock


\bibitem[\protect\citeauthoryear{McFee, Raffel, Liang, Ellis, McVicar,
  Battenberg, and Nieto}{McFee et~al\mbox{.}}{2015}]%
        {mcfee2015librosa}
\bibfield{author}{\bibinfo{person}{Brian McFee}, \bibinfo{person}{Colin
  Raffel}, \bibinfo{person}{Dawen Liang}, \bibinfo{person}{Daniel~P Ellis},
  \bibinfo{person}{Matt McVicar}, \bibinfo{person}{Eric Battenberg}, {and}
  \bibinfo{person}{Oriol Nieto}.} \bibinfo{year}{2015}\natexlab{}.
\newblock \showarticletitle{librosa: Audio and music signal analysis in
  python}. In \bibinfo{booktitle}{\emph{Proceedings of the 14th python in
  science conference}}, Vol.~\bibinfo{volume}{8}. Citeseer,
  \bibinfo{pages}{18--25}.
\newblock


\bibitem[\protect\citeauthoryear{Paszke, Gross, Massa, Lerer, Bradbury, Chanan,
  Killeen, Lin, Gimelshein, Antiga, et~al\mbox{.}}{Paszke
  et~al\mbox{.}}{2019}]%
        {paszke2019pytorch}
\bibfield{author}{\bibinfo{person}{Adam Paszke}, \bibinfo{person}{Sam Gross},
  \bibinfo{person}{Francisco Massa}, \bibinfo{person}{Adam Lerer},
  \bibinfo{person}{James Bradbury}, \bibinfo{person}{Gregory Chanan},
  \bibinfo{person}{Trevor Killeen}, \bibinfo{person}{Zeming Lin},
  \bibinfo{person}{Natalia Gimelshein}, \bibinfo{person}{Luca Antiga},
  {et~al\mbox{.}}} \bibinfo{year}{2019}\natexlab{}.
\newblock \showarticletitle{Pytorch: An imperative style, high-performance deep
  learning library}.
\newblock \bibinfo{journal}{\emph{Advances in neural information processing
  systems}}  \bibinfo{volume}{32} (\bibinfo{year}{2019}).
\newblock


\bibitem[\protect\citeauthoryear{Pennington, Socher, and Manning}{Pennington
  et~al\mbox{.}}{2014}]%
        {pennington2014glove}
\bibfield{author}{\bibinfo{person}{Jeffrey Pennington},
  \bibinfo{person}{Richard Socher}, {and} \bibinfo{person}{Christopher~D
  Manning}.} \bibinfo{year}{2014}\natexlab{}.
\newblock \showarticletitle{Glove: Global vectors for word representation}. In
  \bibinfo{booktitle}{\emph{Proceedings of the 2014 conference on empirical
  methods in natural language processing (EMNLP)}}.
  \bibinfo{pages}{1532--1543}.
\newblock


\bibitem[\protect\citeauthoryear{Raux}{Raux}{2008}]%
        {raux2008flexible}
\bibfield{author}{\bibinfo{person}{Antoine Raux}.}
  \bibinfo{year}{2008}\natexlab{}.
\newblock \showarticletitle{Flexible turn-taking for spoken dialog systems}.
\newblock \bibinfo{journal}{\emph{Language Technologies Institute, CMU Dec}}
  \bibinfo{volume}{12} (\bibinfo{year}{2008}).
\newblock


\bibitem[\protect\citeauthoryear{Raux and Eskenazi}{Raux and Eskenazi}{2009}]%
        {raux2009finite}
\bibfield{author}{\bibinfo{person}{Antoine Raux} {and} \bibinfo{person}{Maxine
  Eskenazi}.} \bibinfo{year}{2009}\natexlab{}.
\newblock \showarticletitle{A finite-state turn-taking model for spoken dialog
  systems}. In \bibinfo{booktitle}{\emph{Proceedings of human language
  technologies: The 2009 annual conference of the North American chapter of the
  association for computational linguistics}}. \bibinfo{pages}{629--637}.
\newblock


\bibitem[\protect\citeauthoryear{Razavi, Kane, and Schubert}{Razavi
  et~al\mbox{.}}{2019}]%
        {razavi2019investigating}
\bibfield{author}{\bibinfo{person}{Seyedeh~Zahra Razavi},
  \bibinfo{person}{Benjamin Kane}, {and} \bibinfo{person}{Lenhart~K Schubert}.}
  \bibinfo{year}{2019}\natexlab{}.
\newblock \showarticletitle{Investigating Linguistic and Semantic Features for
  Turn-Taking Prediction in Open-Domain Human-Computer Conversation.}. In
  \bibinfo{booktitle}{\emph{INTERSPEECH}}. \bibinfo{pages}{4140--4144}.
\newblock


\bibitem[\protect\citeauthoryear{Sacks, Schegloff, and Jefferson}{Sacks
  et~al\mbox{.}}{1978}]%
        {sacks1978simplest}
\bibfield{author}{\bibinfo{person}{Harvey Sacks}, \bibinfo{person}{Emanuel~A
  Schegloff}, {and} \bibinfo{person}{Gail Jefferson}.}
  \bibinfo{year}{1978}\natexlab{}.
\newblock \showarticletitle{A simplest systematics for the organization of turn
  taking for conversation}.
\newblock In \bibinfo{booktitle}{\emph{Studies in the organization of
  conversational interaction}}. \bibinfo{publisher}{Elsevier},
  \bibinfo{pages}{7--55}.
\newblock


\bibitem[\protect\citeauthoryear{Selfridge, Arizmendi, Heeman, and
  Williams}{Selfridge et~al\mbox{.}}{2013}]%
        {selfridge2013continuously}
\bibfield{author}{\bibinfo{person}{Ethan Selfridge}, \bibinfo{person}{Iker
  Arizmendi}, \bibinfo{person}{Peter~A Heeman}, {and} \bibinfo{person}{Jason~D
  Williams}.} \bibinfo{year}{2013}\natexlab{}.
\newblock \showarticletitle{Continuously predicting and processing barge-in
  during a live spoken dialogue task}. In \bibinfo{booktitle}{\emph{Proceedings
  of the SIGDIAL 2013 Conference}}. \bibinfo{pages}{384--393}.
\newblock


\bibitem[\protect\citeauthoryear{Skantze}{Skantze}{2021}]%
        {skantze2021turn}
\bibfield{author}{\bibinfo{person}{Gabriel Skantze}.}
  \bibinfo{year}{2021}\natexlab{}.
\newblock \showarticletitle{Turn-taking in conversational systems and
  human-robot interaction: a review}.
\newblock \bibinfo{journal}{\emph{Computer Speech \& Language}}
  \bibinfo{volume}{67} (\bibinfo{year}{2021}), \bibinfo{pages}{101178}.
\newblock


\bibitem[\protect\citeauthoryear{Snyder, Chen, and Povey}{Snyder
  et~al\mbox{.}}{2015}]%
        {snyder2015musan}
\bibfield{author}{\bibinfo{person}{David Snyder}, \bibinfo{person}{Guoguo
  Chen}, {and} \bibinfo{person}{Daniel Povey}.}
  \bibinfo{year}{2015}\natexlab{}.
\newblock \bibinfo{title}{{MUSAN}: {A} {M}usic, {S}peech, and {N}oise
  {C}orpus}.
\newblock
\newblock
\showeprint{1510.08484}
\newblock
\shownote{arXiv:1510.08484v1.}


\bibitem[\protect\citeauthoryear{Sohn, Berthelot, Carlini, Zhang, Zhang,
  Raffel, Cubuk, Kurakin, and Li}{Sohn et~al\mbox{.}}{2020}]%
        {SohnBCZZRCKL20}
\bibfield{author}{\bibinfo{person}{Kihyuk Sohn}, \bibinfo{person}{David
  Berthelot}, \bibinfo{person}{Nicholas Carlini}, \bibinfo{person}{Zizhao
  Zhang}, \bibinfo{person}{Han Zhang}, \bibinfo{person}{Colin~A Raffel},
  \bibinfo{person}{Ekin~Dogus Cubuk}, \bibinfo{person}{Alexey Kurakin}, {and}
  \bibinfo{person}{Chun-Liang Li}.} \bibinfo{year}{2020}\natexlab{}.
\newblock \showarticletitle{Fixmatch: Simplifying semi-supervised learning with
  consistency and confidence}.
\newblock \bibinfo{journal}{\emph{Advances in Neural Information Processing
  Systems}}  \bibinfo{volume}{33} (\bibinfo{year}{2020}),
  \bibinfo{pages}{596--608}.
\newblock


\bibitem[\protect\citeauthoryear{Tsai, Bai, Liang, Kolter, Morency, and
  Salakhutdinov}{Tsai et~al\mbox{.}}{2019}]%
        {TsaiBLKMS19}
\bibfield{author}{\bibinfo{person}{Yao-Hung~Hubert Tsai},
  \bibinfo{person}{Shaojie Bai}, \bibinfo{person}{Paul~Pu Liang},
  \bibinfo{person}{J~Zico Kolter}, \bibinfo{person}{Louis-Philippe Morency},
  {and} \bibinfo{person}{Ruslan Salakhutdinov}.}
  \bibinfo{year}{2019}\natexlab{}.
\newblock \showarticletitle{Multimodal Transformer for Unaligned Multimodal
  Language Sequences}. In \bibinfo{booktitle}{\emph{Proceedings of the 57th
  Annual Meeting of the Association for Computational Linguistics}}.
  \bibinfo{pages}{6558--6569}.
\newblock


\bibitem[\protect\citeauthoryear{Tsoumakas and Katakis}{Tsoumakas and
  Katakis}{2006}]%
        {tsoumakas2006multi}
\bibfield{author}{\bibinfo{person}{G Tsoumakas} {and} \bibinfo{person}{I
  Katakis}.} \bibinfo{year}{2006}\natexlab{}.
\newblock \showarticletitle{Multi-label classification: An overview
  International Journal of Data Warehousing and Mining}.
\newblock \bibinfo{journal}{\emph{The label powerset algorithm is called PT3}}
  \bibinfo{volume}{3}, \bibinfo{number}{3} (\bibinfo{year}{2006}).
\newblock


\bibitem[\protect\citeauthoryear{Wang, Pan, Liu, Chen, Qiu, Zhou, Huang, Chen,
  Lin, and Cai}{Wang et~al\mbox{.}}{2021a}]%
        {wang2021mell}
\bibfield{author}{\bibinfo{person}{Chengyu Wang}, \bibinfo{person}{Haojie Pan},
  \bibinfo{person}{Yuan Liu}, \bibinfo{person}{Kehan Chen},
  \bibinfo{person}{Minghui Qiu}, \bibinfo{person}{Wei Zhou},
  \bibinfo{person}{Jun Huang}, \bibinfo{person}{Haiqing Chen},
  \bibinfo{person}{Wei Lin}, {and} \bibinfo{person}{Deng Cai}.}
  \bibinfo{year}{2021}\natexlab{a}.
\newblock \showarticletitle{Mell: Large-scale extensible user intent
  classification for dialogue systems with meta lifelong learning}. In
  \bibinfo{booktitle}{\emph{Proceedings of the 27th ACM SIGKDD Conference on
  Knowledge Discovery \& Data Mining}}. \bibinfo{pages}{3649--3659}.
\newblock


\bibitem[\protect\citeauthoryear{Wang, Zheng, Jiang, and Huang}{Wang
  et~al\mbox{.}}{2021b}]%
        {wang2021diversifying}
\bibfield{author}{\bibinfo{person}{Yida Wang}, \bibinfo{person}{Yinhe Zheng},
  \bibinfo{person}{Yong Jiang}, {and} \bibinfo{person}{Minlie Huang}.}
  \bibinfo{year}{2021}\natexlab{b}.
\newblock \showarticletitle{Diversifying Dialog Generation via Adaptive Label
  Smoothing}. In \bibinfo{booktitle}{\emph{Proceedings of the 59th Annual
  Meeting of the Association for Computational Linguistics and the 11th
  International Joint Conference on Natural Language Processing (Volume 1: Long
  Papers)}}. \bibinfo{pages}{3507--3520}.
\newblock


\bibitem[\protect\citeauthoryear{Xie, Dai, Hovy, Luong, and Le}{Xie
  et~al\mbox{.}}{2020}]%
        {XieDHL020}
\bibfield{author}{\bibinfo{person}{Qizhe Xie}, \bibinfo{person}{Zihang Dai},
  \bibinfo{person}{Eduard Hovy}, \bibinfo{person}{Thang Luong}, {and}
  \bibinfo{person}{Quoc Le}.} \bibinfo{year}{2020}\natexlab{}.
\newblock \showarticletitle{Unsupervised data augmentation for consistency
  training}.
\newblock \bibinfo{journal}{\emph{Advances in Neural Information Processing
  Systems}}  \bibinfo{volume}{33} (\bibinfo{year}{2020}),
  \bibinfo{pages}{6256--6268}.
\newblock


\bibitem[\protect\citeauthoryear{Yan and Zhao}{Yan and Zhao}{2018}]%
        {yan2018coupled}
\bibfield{author}{\bibinfo{person}{Rui Yan} {and} \bibinfo{person}{Dongyan
  Zhao}.} \bibinfo{year}{2018}\natexlab{}.
\newblock \showarticletitle{Coupled context modeling for deep chit-chat:
  towards conversations between human and computer}. In
  \bibinfo{booktitle}{\emph{Proceedings of the 24th ACM SIGKDD International
  Conference on Knowledge Discovery \& Data Mining}}.
  \bibinfo{pages}{2574--2583}.
\newblock


\bibitem[\protect\citeauthoryear{Yang, Wang, Yi, Zhu, Rehman, Zadeh, Poria, and
  Morency}{Yang et~al\mbox{.}}{2021}]%
        {YangWYZRZPM21}
\bibfield{author}{\bibinfo{person}{Jianing Yang}, \bibinfo{person}{Yongxin
  Wang}, \bibinfo{person}{Ruitao Yi}, \bibinfo{person}{Yuying Zhu},
  \bibinfo{person}{Azaan Rehman}, \bibinfo{person}{Amir Zadeh},
  \bibinfo{person}{Soujanya Poria}, {and} \bibinfo{person}{Louis-Philippe
  Morency}.} \bibinfo{year}{2021}\natexlab{}.
\newblock \showarticletitle{MTAG: Modal-Temporal Attention Graph for Unaligned
  Human Multimodal Language Sequences}. In
  \bibinfo{booktitle}{\emph{NAACL-HLT}}.
\newblock


\bibitem[\protect\citeauthoryear{Yngve}{Yngve}{1970}]%
        {yngve1970getting}
\bibfield{author}{\bibinfo{person}{Victor~H Yngve}.}
  \bibinfo{year}{1970}\natexlab{}.
\newblock \showarticletitle{On getting a word in edgewise}. In
  \bibinfo{booktitle}{\emph{Chicago Linguistics Society, 6th Meeting, 1970}}.
  \bibinfo{pages}{567--578}.
\newblock


\bibitem[\protect\citeauthoryear{Yu, Xu, Yuan, and Wu}{Yu
  et~al\mbox{.}}{2021}]%
        {yu2021learning}
\bibfield{author}{\bibinfo{person}{Wenmeng Yu}, \bibinfo{person}{Hua Xu},
  \bibinfo{person}{Ziqi Yuan}, {and} \bibinfo{person}{Jiele Wu}.}
  \bibinfo{year}{2021}\natexlab{}.
\newblock \showarticletitle{Learning Modality-Specific Representations with
  Self-Supervised Multi-Task Learning for Multimodal Sentiment Analysis}. In
  \bibinfo{booktitle}{\emph{Proceedings of the AAAI Conference on Artificial
  Intelligence}}, Vol.~\bibinfo{volume}{35}. \bibinfo{pages}{10790--10797}.
\newblock


\bibitem[\protect\citeauthoryear{Zadeh, Chen, Poria, Cambria, and
  Morency}{Zadeh et~al\mbox{.}}{2017}]%
        {zadeh-etal-2017-tensor}
\bibfield{author}{\bibinfo{person}{Amir Zadeh}, \bibinfo{person}{Minghai Chen},
  \bibinfo{person}{Soujanya Poria}, \bibinfo{person}{Erik Cambria}, {and}
  \bibinfo{person}{Louis-Philippe Morency}.} \bibinfo{year}{2017}\natexlab{}.
\newblock \showarticletitle{Tensor Fusion Network for Multimodal Sentiment
  Analysis}. In \bibinfo{booktitle}{\emph{Proceedings of the 2017 Conference on
  Empirical Methods in Natural Language Processing}}.
  \bibinfo{pages}{1103--1114}.
\newblock


\bibitem[\protect\citeauthoryear{Zhang, Li, Xu, Zhang, Zhao, and Gao}{Zhang
  et~al\mbox{.}}{2021a}]%
        {zhang-etal-2021-textoir}
\bibfield{author}{\bibinfo{person}{Hanlei Zhang}, \bibinfo{person}{Xiaoteng
  Li}, \bibinfo{person}{Hua Xu}, \bibinfo{person}{Panpan Zhang},
  \bibinfo{person}{Kang Zhao}, {and} \bibinfo{person}{Kai Gao}.}
  \bibinfo{year}{2021}\natexlab{a}.
\newblock \showarticletitle{{TEXTOIR}: An Integrated and Visualized Platform
  for Text Open Intent Recognition}. In \bibinfo{booktitle}{\emph{Proceedings
  of the 59th Annual Meeting of the Association for Computational Linguistics
  and the 11th International Joint Conference on Natural Language Processing:
  System Demonstrations}}. \bibinfo{pages}{167--174}.
\newblock


\bibitem[\protect\citeauthoryear{Zhang, Xu, and Lin}{Zhang
  et~al\mbox{.}}{2021b}]%
        {Zhang_Xu_Lin_2021}
\bibfield{author}{\bibinfo{person}{Hanlei Zhang}, \bibinfo{person}{Hua Xu},
  {and} \bibinfo{person}{Ting-En Lin}.} \bibinfo{year}{2021}\natexlab{b}.
\newblock \showarticletitle{Deep Open Intent Classification with Adaptive
  Decision Boundary}.
\newblock \bibinfo{journal}{\emph{Proceedings of the AAAI Conference on
  Artificial Intelligence}} \bibinfo{volume}{35}, \bibinfo{number}{16}
  (\bibinfo{date}{May} \bibinfo{year}{2021}), \bibinfo{pages}{14374--14382}.
\newblock


\bibitem[\protect\citeauthoryear{Zhang, Xu, Lin, and Lyu}{Zhang
  et~al\mbox{.}}{2021c}]%
        {Zhang_Xu_Lin_Lyu_2021}
\bibfield{author}{\bibinfo{person}{Hanlei Zhang}, \bibinfo{person}{Hua Xu},
  \bibinfo{person}{Ting-En Lin}, {and} \bibinfo{person}{Rui Lyu}.}
  \bibinfo{year}{2021}\natexlab{c}.
\newblock \showarticletitle{Discovering New Intents with Deep Aligned
  Clustering}.
\newblock \bibinfo{journal}{\emph{Proceedings of the AAAI Conference on
  Artificial Intelligence}} \bibinfo{volume}{35}, \bibinfo{number}{16}
  (\bibinfo{year}{2021}), \bibinfo{pages}{14365--14373}.
\newblock


\bibitem[\protect\citeauthoryear{Zhang, Zheng, Shao, Mao, Xi, and Huang}{Zhang
  et~al\mbox{.}}{2020}]%
        {zhang2020dialogue}
\bibfield{author}{\bibinfo{person}{Rongsheng Zhang}, \bibinfo{person}{Yinhe
  Zheng}, \bibinfo{person}{Jianzhi Shao}, \bibinfo{person}{Xiaoxi Mao},
  \bibinfo{person}{Yadong Xi}, {and} \bibinfo{person}{Minlie Huang}.}
  \bibinfo{year}{2020}\natexlab{}.
\newblock \showarticletitle{Dialogue Distillation: Open-Domain Dialogue
  Augmentation Using Unpaired Data}. In \bibinfo{booktitle}{\emph{Proceedings
  of the 2020 Conference on Empirical Methods in Natural Language Processing
  (EMNLP)}}. \bibinfo{pages}{3449--3460}.
\newblock


\bibitem[\protect\citeauthoryear{Zhang, Hu, Wu, Wu, Li, Sun, Yuan, and
  Wang}{Zhang et~al\mbox{.}}{2022}]%
        {zhang-etal-2022-slot}
\bibfield{author}{\bibinfo{person}{Sai Zhang}, \bibinfo{person}{Yuwei Hu},
  \bibinfo{person}{Yuchuan Wu}, \bibinfo{person}{Jiaman Wu},
  \bibinfo{person}{Yongbin Li}, \bibinfo{person}{Jian Sun},
  \bibinfo{person}{Caixia Yuan}, {and} \bibinfo{person}{Xiaojie Wang}.}
  \bibinfo{year}{2022}\natexlab{}.
\newblock \showarticletitle{A Slot Is Not Built in One Utterance: Spoken
  Language Dialogs with Sub-Slots}. In \bibinfo{booktitle}{\emph{Findings of
  the Association for Computational Linguistics: ACL 2022}}.
  \bibinfo{publisher}{Association for Computational Linguistics},
  \bibinfo{address}{Dublin, Ireland}, \bibinfo{pages}{309--321}.
\newblock


\bibitem[\protect\citeauthoryear{Zhao, Black, and Eskenazi}{Zhao
  et~al\mbox{.}}{2015}]%
        {zhao2015incremental}
\bibfield{author}{\bibinfo{person}{Tiancheng Zhao}, \bibinfo{person}{Alan~W
  Black}, {and} \bibinfo{person}{Maxine Eskenazi}.}
  \bibinfo{year}{2015}\natexlab{}.
\newblock \showarticletitle{An incremental turn-taking model with active system
  barge-in for spoken dialog systems}. In \bibinfo{booktitle}{\emph{Proceedings
  of the 16th Annual Meeting of the Special Interest Group on Discourse and
  Dialogue}}. \bibinfo{pages}{42--50}.
\newblock


\bibitem[\protect\citeauthoryear{Zhao, Tian, Yao, Zheng, Lee, Song, Sun, and
  Zhang}{Zhao et~al\mbox{.}}{2022}]%
        {zhao-etal-2022-improving}
\bibfield{author}{\bibinfo{person}{Yingxiu Zhao}, \bibinfo{person}{Zhiliang
  Tian}, \bibinfo{person}{Huaxiu Yao}, \bibinfo{person}{Yinhe Zheng},
  \bibinfo{person}{Dongkyu Lee}, \bibinfo{person}{Yiping Song},
  \bibinfo{person}{Jian Sun}, {and} \bibinfo{person}{Nevin Zhang}.}
  \bibinfo{year}{2022}\natexlab{}.
\newblock \showarticletitle{Improving Meta-learning for Low-resource Text
  Classification and Generation via Memory Imitation}. In
  \bibinfo{booktitle}{\emph{Proceedings of the 60th Annual Meeting of the
  Association for Computational Linguistics (Volume 1: Long Papers)}}.
  \bibinfo{publisher}{Association for Computational Linguistics},
  \bibinfo{address}{Dublin, Ireland}, \bibinfo{pages}{583--595}.
\newblock


\bibitem[\protect\citeauthoryear{Zheng, Chen, Liu, and Sun}{Zheng
  et~al\mbox{.}}{2021}]%
        {zheng2021mmchat}
\bibfield{author}{\bibinfo{person}{Yinhe Zheng}, \bibinfo{person}{Guanyi Chen},
  \bibinfo{person}{Xin Liu}, {and} \bibinfo{person}{jian Sun}.}
  \bibinfo{year}{2021}\natexlab{}.
\newblock \showarticletitle{MMChat: Multi-Modal Chat Dataset on Social Media}.
\newblock \bibinfo{journal}{\emph{arXiv preprint arXiv:2108.07154}}
  (\bibinfo{year}{2021}).
\newblock


\bibitem[\protect\citeauthoryear{Zheng, Zhang, Huang, and Mao}{Zheng
  et~al\mbox{.}}{2020}]%
        {zheng2020pre}
\bibfield{author}{\bibinfo{person}{Yinhe Zheng}, \bibinfo{person}{Rongsheng
  Zhang}, \bibinfo{person}{Minlie Huang}, {and} \bibinfo{person}{Xiaoxi Mao}.}
  \bibinfo{year}{2020}\natexlab{}.
\newblock \showarticletitle{A pre-training based personalized dialogue
  generation model with persona-sparse data}. In
  \bibinfo{booktitle}{\emph{Proceedings of the AAAI Conference on Artificial
  Intelligence}}, Vol.~\bibinfo{volume}{34}. \bibinfo{pages}{9693--9700}.
\newblock


\bibitem[\protect\citeauthoryear{Zhou, Ke, Zhang, Gu, Zheng, Zheng, Wang, Wu,
  Sun, Yang, et~al\mbox{.}}{Zhou et~al\mbox{.}}{2021}]%
        {zhou2021eva}
\bibfield{author}{\bibinfo{person}{Hao Zhou}, \bibinfo{person}{Pei Ke},
  \bibinfo{person}{Zheng Zhang}, \bibinfo{person}{Yuxian Gu},
  \bibinfo{person}{Yinhe Zheng}, \bibinfo{person}{Chujie Zheng},
  \bibinfo{person}{Yida Wang}, \bibinfo{person}{Chen~Henry Wu},
  \bibinfo{person}{Hao Sun}, \bibinfo{person}{Xiaocong Yang}, {et~al\mbox{.}}}
  \bibinfo{year}{2021}\natexlab{}.
\newblock \showarticletitle{EVA: An open-domain chinese dialogue system with
  large-scale generative pre-training}.
\newblock \bibinfo{journal}{\emph{arXiv preprint arXiv:2108.01547}}
  (\bibinfo{year}{2021}).
\newblock


\bibitem[\protect\citeauthoryear{Zhou, Gao, Li, and Shum}{Zhou
  et~al\mbox{.}}{2020}]%
        {zhou2020design}
\bibfield{author}{\bibinfo{person}{Li Zhou}, \bibinfo{person}{Jianfeng Gao},
  \bibinfo{person}{Di Li}, {and} \bibinfo{person}{Heung-Yeung Shum}.}
  \bibinfo{year}{2020}\natexlab{}.
\newblock \showarticletitle{The design and implementation of xiaoice, an
  empathetic social chatbot}.
\newblock \bibinfo{journal}{\emph{Computational Linguistics}}
  \bibinfo{volume}{46}, \bibinfo{number}{1} (\bibinfo{year}{2020}),
  \bibinfo{pages}{53--93}.
\newblock


\end{thebibliography}

\appendix

\section{Appendix}
\subsection{Experimental Settings}
We empirical set the hyperparameter $\alpha$ as 0.25 for multimodal data augmentation. Note that when $\alpha = 1$, the beta distribution becomes a uniform distribution. We initialize our word embedding with the pre-trained GloVe \cite{pennington2014glove}. The training batch size is 256, and the learning rate for user state detection, backchannel, and barge-in is 4e-4, 2e-4, and 5e-4, respectively. All models are implemented in PyTorch \cite{paszke2019pytorch}. We use the audios in WAV format with 8k sample rates. We use Librosa \cite{mcfee2015librosa} to transform PCM data in WAV files and audio streams into log Mel-filterbank (FBANK) as input of the audio encoder. We set the number of mel bands, length of the FFT window, and the number of samples between successive frames as 64, 1024, 512, respectively.

\subsection{Hamming Loss}
Given that $\hat{y}_i$ is the predicted values of the i-th label of the sample, we define $\mathcal{L}_\text{{Hamming}}$ as follow: 
\begin{align}
    \mathcal{L}_\text{{Hamming}}(y, \hat{y}) = \frac{1}{n_\text{labels}} \sum_{i=0}^{n_\text{labels}-1} 1(\hat{y_i} \neq y_i) 
\end{align}
where 1(x) is the indicator function. 

\subsection{Selection of Evaluation Metrics}
The reason we choose precision and recall instead of accuracy for barge-in is twofold: on the one hand, since most requests are not barge-in, accuracy could not truly reflect the performance of the model. If the model identifies all requests as not barge-in, it may have high accuracy despite many false-negative requests. On the other hand, precision is more important than recall because false positives can interrupt agents and result in a poor experience. In contrast, false negatives still have a chance to be correctly identified during continuous prediction.

\subsection{Design Choice for Feature Extraction}
The reason we choose simple models for feature extraction modules is twofold. We found that using complex architecture, such as replacing GRU with bidirectional LSTM or vanilla transformer, does not achieve significant improvements compared with the simple one. Another is latency and computational cost. Since there is no significant difference in accuracy, using a simple model can result in faster runtime, lower CPU usage, and cost savings.

\end{document}